\pdfoutput=1
\documentclass{article}

\usepackage{preprint,times}

\usepackage[utf8]{inputenc}
\usepackage[T1]{fontenc}
\usepackage{xcolor}
\definecolor{darkgreen}{RGB}{20,100,20}
\usepackage[
    colorlinks=true,
    linkcolor=teal,
    filecolor=magenta,
    urlcolor=teal,
    citecolor=teal,
    pdftitle={Mechanistic Evidence for Spectral Structures in Prior-Data Fitted Networks},
    pdfauthor={Kaustubh Sharma, Srijan Tiwari, Ojasva Nema, Parikshit Pareek}
]{hyperref}
\usepackage{url}
\newcommand{\Hbar}{\mkern 1.5mu\overline{\mkern-1.5mu H\mkern-1.5mu}\mkern 1.5mu}
\usepackage{booktabs}
\usepackage{amsfonts}
\usepackage{amsmath}
\usepackage{amssymb}
\usepackage{nicefrac}
\usepackage{microtype}
\usepackage{xcolor}
\usepackage{graphicx}

\usepackage{subcaption}
\usepackage{multirow}
\usepackage{enumitem}
\usepackage{mathtools}
\usepackage{amsthm}
\usepackage{tikz}
\usepackage{float}
\usetikzlibrary{positioning,fit,arrows.meta,calc,backgrounds}

\newtheorem{theorem}{Theorem}
\newtheorem{proposition}{Proposition}

\title{Mechanistic Evidence for Spectral Structures in Prior-Data Fitted Networks}

\author{
  Kaustubh Sharma$^\star$, Srijan Tiwari$^\star$, Ojasva Nema \& Parikshit Pareek$^\dagger$ \\
  Indian Institute of Technology Roorkee, India \\
  \texttt{pareek@ee.iitr.ac.in}
}

\begin{document}

\maketitle
{\let\thefootnote\relax\footnotetext{%
  $^\star$Equal contribution. \quad $^\dagger$Corresponding author.\\
  This work was supported by ANRF PM Early Career Research Grant (ANRF/ECRG/2024/001962/ENS) and the IIT Roorkee Faculty Initiation Grant (IITR/SRIC/1431/FIG-101078). Authors also acknowledge Avathon Physical AI Lab, IIT Roorkee funding support from Avathon Inc. USA.}}

\begin{abstract}
Prior-Data Fitted Networks (PFNs) perform approximate Bayesian inference in a single forward pass, and tabular foundation models (TFMs) built on them are now widely used. To understand what networks infer internally, recent mechanistic studies of TFMs locate where predictions form, but treat these models as tabular predictors rather than as PFNs. It therefore remains unknown whether PFNs represent the spectral content of their context, the quantity that specifies a stationary kernel, and whether this content can be read out as an explicit kernel. We answer both questions. First, across seven PFNs, including four pretrained TFMs and a model trained only on a decision-tree prior, a linear probe on the residual stream recovers the frequency of the context with $R^2 \geq 0.95$. This structure is led by a single principal direction. Second, activation and subspace patching show that the network uses the structure through a low-dimensional subspace, where a few spectral directions move predictions far more than random ones. This holds even for the decision-tree model, so a spectral training prior is not required. On real datasets with up to 499 features, 64 of the 192 directions of TabPFN, chosen without labels, carry 85 to 95\% of the causal effect of the context in all but one pair. Third, we introduce a Filter Bank Decoder that turns frozen PFN representations into an explicit stationary kernel through Bochner's theorem. Without any test-time optimization, the decoded kernel supports Gaussian process regression competitive with deep kernel learning and random Fourier features at about $200\times$ lower cost. PFN latents therefore hold spectral structure that is causally used and recoverable as a portable kernel.
\end{abstract}

\section{Introduction} \label{sec:intro}

Prior-Data Fitted Networks (PFNs) offer a resolution to the computational bottleneck of Bayesian inference through amortization \citep{muller2022transformers}. A PFN is trained on a distribution of synthetic tasks sampled from a prior, learning to map context data directly to posterior predictive distributions (PPDs) in a single forward pass. This enables efficient inference on new datasets drawn from the same prior family, without per-dataset fitting or optimization. Building on PFNs, tabular foundation models (TFMs) have fulfilled the long-standing need for structured-data foundation models and are now state of the art on tabular benchmarks \citep{hollmann2025accurate,qu2026tabiclv2}.Throughout, we use ``PFN'' for any transformer trained on synthetic datasets drawn from a prior to predict in context which includes TFMs.

However, this efficiency comes at the cost of \textbf{spectral structure opacity}. In methods like Gaussian Process (GP), the kernel is an explicit, interpretable, portable object that can be inspected, composed, and transferred to new tasks \citep{duvenaud2013structurediscoverynonparametricregression}. In a PFN, the corresponding structure is absorbed implicitly into the network's weights and activations. %
More importantly, the spectral opacity now extends to widely deployed models. TFMs such as TabPFN \citep{hollmann2025accurate,grinsztajn2025tabpfn}, TabICLv2 \citep{qu2026tabiclv2} and TabSwift \citep{liu2026tabswift} are PFNs pretrained only on synthetic tabular priors, which serve as general-purpose predictors on real tables. For these models too, it is not known whether the hidden Bayesian object bears any interpretable relation to the context. 

Mechanistic methods, developed mainly for language models \citep{  belinkov2022probing}, have since been applied to non-language models \citep{el2025mechanisticinterpretabilitygraphtransformers}. Recently, for TFMs, these studies show where predictions form across layers \citep{balef2026onelayer}, which readout rules combine context labels \citep{bilos2026mechanistic}, whether single neurons are concept-selective \citep{knauer2026grandmother}, and how TabPFN's effective bandwidth grows with context size \citep{zheng2025tables}. Preliminary studies also probe task quantities and attention heads in TabPFN, and report encodings that do not carry over across fits or samples \citep{gupta2026lookingglass, gupta2026wherecomputation}. However, three main gaps remain. First, the targets studied are the model's output, its components, or quantities specific to one task family. None of these quantities are something which a Bayesian method would infer from the context, such as the spectral content that fixes a stationary kernel. Second, no study tests whether a TFM encodes a quantity (e.g. frequency), inferred only from the context, in the same way across datasets, and whether the network relies on that encoding for its prediction. Third, none of these analyses turns an internal representation into an object usable outside the network. Although, theoretical constructions show that transformers can implement GP posterior prediction when the kernel is supplied \citep{kang2026ppd, yan2026richardson}, yet kernel is not read from a trained PFN. %
Most of these studies treat TFMs as tabular predictors rather than as PFNs, the amortized Bayesian predictors on which they are built \citep{muller2022transformers}. We instead frame our analysis for PFN-style networks in general, and apply the same probes and interventions to small PFNs trained on GP priors and to large pretrained TFMs alike, establishing generality of analysis.

These gaps are instances of two open challenges that \citet{muller2025position} identify for PFNs to serve as general-purpose Bayesian tools: \textbf{1) the lack of mechanistic transparency}, since we do not know \emph{where} or \emph{how} Bayesian structure is represented inside the network, nor whether it plays a causal role in prediction; and \textbf{2) the absence of explicit, portable representations} of the priors learned during amortized inference. We address these by asking two complementary questions: \textbf{(a)}~\textit{Do PFNs encode the spectral content of their context in a shared, interpretable form that is causally used in prediction?} and \textbf{(b)}~\textit{Can this structure be extracted as an explicit kernel?} Our contributions are:

\textbullet~ \textbf{Spectral structure (question (a)):} We study total seven PFNs: Four TFMs are pretrained on synthetic tabular priors and used as released, two PFNs we train on spectral mixtures (VA-PFN \citep{muller2022transformers} and DVA-PFN \citep{sharma2025dva}), and a nanoTabPFN we train only on decision-tree prior, whose piecewise-constant targets carry no GP or Fourier component. In all seven, a ladder of probes of increasing capacity shows that spectral information, in terms of frequency,  is \textit{present and linearly decodable} from the mean-pooled residual stream of the transformer blocks ($\Hbar$), with a single probe reading it across contexts, and that it is led by one principal direction. In the four PFNs we intervene on (VA-PFN, DVA-PFN, TabPFN 2.5 and nanoTabPFN), activation and subspace patching show that the network \textit{uses this information} for prediction through a low-dimensional subspace, where a few spectral directions move predictions far more than an equal number of random ones. Even under the decision-tree prior, a probe on the six-block nanoTabPFN regressor reaches $R^2 = 0.99$, and patching its 16 most frequency-correlated of 192 directions gives a causal effect (1 = full transfer) of $0.68$, against $0.06$ for 16 random directions. On TabPFN 2.5, the organization also holds for GP draws in 1 to 10 input dimensions. On eight pairs of real datasets with 5 to 499 features, 64 of 192 directions, chosen without labels, carry 85 to 95\% of the context's causal effect.

\textbullet~  \textbf{Kernel read-out for spectral data (question (b)):} We introduce a \emph{Filter Bank Decoder} that maps frozen DVA-PFN representations to explicit spectral densities and, via Bochner's theorem, to closed-form stationary kernels. We show that from a single realization the spectral weights are identifiable at best up to a common scale, which we recover analytically. Peak locations and bandwidths remain recoverable. Controls confirm that the decoder reads the kernel from $\Hbar$ because on latents of a randomly initialized PFN (or with $\Hbar$ removed) its kernel error rises by about two orders of magnitude. The decoded kernels support GP regression competitive with per-task-optimized baselines, namely deep kernel learning and random Fourier features \citep{wilson2015deepkernellearning, NIPS2007_013a006f} while requiring only a single forward pass, running about 200× faster, and giving the best, predictive log-density among these learned methods. On held-out points they also achieve $17$--$78\times$ lower error than spectral mixtures fitted to classical periodograms.

\section{Background and Related Work}
\label{sec:background}

\subsection{Prior-Data Fitted Networks}
\label{sec:pfn}
In \citet{muller2022transformers}, amortized Bayesian inference was enabled by training a neural set-predictor on a vast distribution of synthetic datasets. Formally, given a prior $p(\mathcal{D})$ over supervised learning tasks, we sample datasets $\mathcal{D}_k = \{(x_i, y_i)\}_{i=1}^N \sim p(\mathcal{D})$. Each dataset is partitioned into a context set $\mathcal{D}_{\text{ctx}} = (X_{\text{ctx}}, Y_{\text{ctx}})$ containing observed pairs, and a query set $\mathcal{D}_{\text{q}} = (X_{\text{q}}, Y_{\text{q}})$ containing targets to predict. The model parameters $\theta$ are optimized to minimize the expected Negative Log-Likelihood lo       ss.
PFNs have since been trained on many priors for specific tasks \citep{muller2025position}. %

\textbf{Models studied.} The seven PFNs we used for analysis differ in attention rule and training prior. VA-PFN \citep{muller2022transformers} forms queries, keys and values from $(x,y)$ jointly while DVA-PFN \citep{sharma2025dva} computes attention weights from inputs alone and passes targets only through the value stream. We pretrain both on spectral-mixture GP priors (Appendix~\ref{app:models}).  TabPFN 2.5 (hereafter TabPFN) and three other TFMs, TabPFN 3.0, TabICLv2 and TabSwift \citep{grinsztajn2026tabpfn3, qu2026tabiclv2, liu2026tabswift}, are used as released. The decision-tree nanoTabPFN \citep{pfefferle2025nanotabpfn} is described in Section~\ref{sec:probing}. All PFNs are frozen during analysis.

\textbf{Notation.} A PFN has $L$ transformer blocks of width $d$, each applying attention and then a feed-forward layer, both with a residual connection. $H^{(j)}\in\mathbb{R}^{N\times d}$ is the residual stream after block $j$ at the $N$ query positions, with $H=H^{(L)}$ the latent the prediction head reads, and $\overline{H}=\frac{1}{N}\sum_i H[i]\in\mathbb{R}^d$ its mean over positions (final block unless stated otherwise). $V=\phi_y(Y_{\mathrm{ctx}})$ and $K=\phi_x(X_{\mathrm{ctx}})$ are the value- and key-stream encoding of the context, and $\overline{V}$ is the pooled $V$. Appendix ~\ref{app:models} gives the model-specific details.

\subsection{Spectral Representation of Stationary Kernels}
\label{sec:bochner}
For a stationary covariance kernel $k(\tau)$, Bochner's theorem \citep{bochner1959lectures} establishes a one-to-one correspondence between the kernel and a non-negative spectral density $S(\omega)$ via the Fourier transform: $ k(\tau) = \int_{\mathbb{R}} S(\omega)\, e^{2\pi i \omega \tau}\, d\omega, \, S(\omega) \geq 0.$ This duality means that specifying a kernel is equivalent to specifying a spectral density: smoothness, periodicity, and long-range correlation structure are all encoded in the shape  of $S(\omega)$. Building on this correspondence, \cite{wilson2013gaussian} introduced the 
\emph{Spectral Mixture} (SM) kernel, which models the spectral density as a mixture of $Q$ Gaussians: $S(\omega)=\sum_{q=1}^{Q}\frac{w_q}{2}\Big[\mathcal{N}\!\left(\omega \mid \mu_q,\sigma_q^2\right)+\mathcal{N}\!\left(\omega \mid -\mu_q,\sigma_q^2\right)\Big],\; w_q>0$, where each component is characterized by a center frequency $\mu_q$, a bandwidth $\sigma_q$, and a mixture weight $w_q$. Substituting into Bochner's theorem yields the closed-form kernel $ k(\tau) = \sum_{q=1}^{Q} w_q 
    \exp\!\left(-2\pi^2 \sigma_q^2 \tau^2\right) 
    \cos\!\left(2\pi \mu_q \tau\right).$
The SM kernel is a universal approximator in the space of stationary kernels 
\citep{wilson2013gaussian}: any continuous stationary kernel on a compact domain can be approximated arbitrarily well by choosing a sufficient number of mixture components $Q$.

\subsection{Related Work}\label{sec:mechinterp_background}

\textbf{Mechanistic tools.} Mechanistic interpretability reverse-engineers the computations of trained networks by identifying features and circuits \citep{olah2020zoom}. We use two of its tools. Probing classifiers test whether a quantity is decodable from a representation. We follow the probe ladder of \citet{alain2016understanding} and the control protocol of \citet{hewitt2019designing}, since a sufficiently expressive probe can learn the target itself \citep{belinkov2022probing}. Activation patching tests whether a decodable quantity is used, by replacing activations and measuring the change in output \citep{meng2022locating, vig2020investigating, geiger2021causal}. Distributed alignment search \citep{geiger2024finding} learns the subspace to patch by optimization. We instead select it by ranking principal components by their correlation with the target.

\textbf{Interpretability of tabular foundation models.} \citet{balef2026onelayer} apply a tabular tuned lens to six TFMs and find that predictions are decodable from early layers and that later layers are largely redundant for that purpose. A single looped layer with 20\% of the parameters performs comparably. \citet{bilos2026mechanistic} use causal interventions to identify distinct similarity-based readouts in TabPFNv2, TabICLv2 and Mitra, ranging from an attention-weighted vote over context labels to a class-conditional mean. They trace the models' invariances to positional parameters. A similarity-based readout acts as a kernel smoother, but which similarity is used, and how it is set from the context, is left open. At the level of single units, \citet{knauer2026grandmother, knauer2026concepttracer} find that only a small fraction of neurons are selective for human-defined concepts. \citet{hu2026topological} relate the persistent homology of layer representations to prediction reliability. Preliminary studies probe TabPFN for intermediate quantities such as regression coefficients and localize its computation to a few attention heads \citep{gupta2026lookingglass, gupta2026wherecomputation}. These works localize computation and describe readouts. We instead target one quantity the model must infer from the context, test whether it is encoded in the same way across contexts and used for prediction, and decode it. Complementary work shows that the synthetic training prior shapes what TFMs learn \citep{bouadi2026shaping, zhao2026structural}. Our decision-tree experiment (Section \ref{sec:probing}) adds that the spectral structure emerges even when the prior has no GP or Fourier component.
Further, \citet{zheng2025tables} study TabPFN through signal reconstruction. They define a context kernel, the Jacobian of query predictions with respect to context labels, which reduces to the attention matrix for a single attention block. They show that the range of frequencies TabPFN recovers grows with the context size and is modulated by the positional encoding. This characterizes the function TabPFN computes. We instead locate a representation of the context's frequency content inside the network (Section \ref{sec:locate}), test whether it is used  (Section \ref{sec:causal}), and read it as an explicit kernel (Section \ref{sec:decoder}).

\textbf{Kernel discovery.} Most kernel design methods search or optimize the kernel per dataset \citep{wilson2013gaussian, duvenaud2013structurediscoverynonparametricregression, lloyd2014automaticconstructionnaturallanguagedescription, wilson2015deepkernellearning, tancik2020fourierfeaturesletnetworks}. Amortized methods instead train a dedicated network to map a dataset to spectral-mixture parameters \citep{liu2020taskagnostic}, a kernel structure \citep{simpson2021kitt}, or the hyperparameters of a given structured kernel \citep{bitzer2023amortized}. Theoretical constructions further show that transformers can implement GP posterior prediction in context when the kernel is supplied \citep{kang2026ppd, yan2026richardson}. Our decoder instead reads, from a frozen predictor never trained to output one, the kernel that existing works take as given.

\section{Locating Spectral Structure in PFN Representations}
\label{sec:locate}
First we ask how a trained PFN organizes spectral information about its input. To answer this, we employ \textit{a ladder of probes of increasing capacity} \citep{alain2016understanding, belinkov2022probing}, from a parameter-free geometric test to linear and then nonlinear read-outs. Each step adds parameters and so tests a weaker notion of accessibility, making agreement at the lower stages the stronger evidence.
First we probe six PFNs below and then the tree-prior-based nanoTabPFN in Section~\ref{sec:probing}. Causal tests follow in Section~\ref{sec:causal}, and Appendix~\ref{app:spectral} gives protocols and supporting results for this section.

\subsection{Parameter-Free Probing: Frequency-Driven Geometry of \texorpdfstring{$\Hbar$}{H}}
\label{sec:manifold} 
The most conservative probe has zero trainable parameters. We ask a simple question: when the generating frequency $f$ of an input sinusoid changes, does $\overline{H}$ change in a correspondingly structured way? We generate 500 sinusoids $y(t)=\sin(2\pi f_i t+\varphi)$ with $f_i\sim\mathcal{U}[0.5,5.0]$\,Hz and random phase $\varphi\sim\mathcal{U}[0,2\pi]$, pass each through the frozen PFN, and read $\overline{H}$ (Notation, Section~\ref{sec:pfn}). In DVA-PFN, where $\phi_x$ is a Fourier-feature encoder and $\phi_y$ an MLP, the latent starts as $H^{(0)}=\phi_x(X_q)$ and block $j$ updates it as $H^{(j)}=\mathrm{Block}_j\big(H^{(j-1)};\phi_x(X_{\mathrm{ctx}}),\phi_y(Y_{\mathrm{ctx}})\big)$, with queries from $H^{(j-1)}$, keys from $\phi_x(X_{\mathrm{ctx}})$ and values from $\phi_y(Y_{\mathrm{ctx}})$. More details are in Appendix~\ref{app:models}.

First, \textit{is the global geometry of $\Hbar$ governed by $f$?} We calculate the Pearson correlation between two vectors of pairwise distances as $\rho_{\Delta} \;=\; \mathrm{Corr}\!\left(\{\Delta f_{ij}\},\,\{\Delta H_{ij}\}\right)$, which measure change in latent embedding ($\Delta H_{ij}=\|\Hbar_i-\Hbar_j\|_2$) with change in input frequency ($\Delta f_{ij}=|f_i - f_j|$). This measures whether frequency-close signals are embedding-close.
Second, we ask \textit{if the frequency sensitivity is dominant in a particular direction in $\Hbar$?} For this, we report the Pearson correlation $|r|_{\text{PC}_0}$ between the generating frequency and $\Hbar$ projected onto its first principal component in the original $d$-dimensional space. Let $\mathrm{PC}_0=\arg\max_{\lVert v\rVert=1} v^\top\Sigma v$ with $\Sigma=\frac{1}{n}\sum_{i=1}^{n}(\overline{H}_i-\mathbf{m})(\overline{H}_i-\mathbf{m})^\top$, where $\mathbf{m}$ is the mean of $\overline{H}$ over the $n$ signals. Then $|r|_{\mathrm{PC}_0}=\bigl|\mathrm{Corr}(\{z_i\}_{i=1}^{n},\{f_i\}_{i=1}^{n})\bigr|$ with $z_i=\mathrm{PC}_0^\top(\overline{H}_i-\mathbf{m})$. This isolates the dominant mode of variation rather than the full geometry. Table~\ref{tab:parameter-free} reports both these metrics. In TabPFN, $\Hbar$ tracks frequency at $\rho_{\Delta}\!=\!0.765$ and $|r|_{\text{PC}_0}\!=\!0.882$, and the three further TFMs do so more strongly ($\rho_{\Delta}$ from 0.79 to 0.94). %

\textbf{Where Does Spectral Information Live?}
Repeating the parameter-free measurement on $\overline{V}$ gives a sharp architectural split: $\rho_V\!=\!0.19$ (DVA), $0.53$ (VA), $0.79$ (TabPFN). This matches how each mechanism handles the value stream: DVA-PFN routes $y$-information to $\overline{V}$ alone and confines frequency to $\Hbar$; VA lets the joint $(x,y)$ embedding mix into all three projections, so frequency leaks into $\overline{V}$; and TabPFN's alternating row/column attention repeatedly updates value representations alongside feature context, producing the strongest leakage. Thus, we read the $\Hbar$--$\overline{V}$ split as architectural evidence that attention design governs the \emph{localization} only not the \emph{existence}. %

\subsection{Linear Accessibility and Its Controls}
\label{sec:probing}

Last subsections, geometric alignment shows that $\Hbar$ is \emph{organized} by frequency, not that spectral quantities can be \emph{read out} simply. We therefore train a linear probe (a lower bound on accessibility) and an MLP probe (a higher-capacity check) on the same frozen $\Hbar$. If both reach high $R^2$ and agree, the target is encoded as an approximately linear function of $\Hbar$.

\textbf{A linear read-out is enough.} For two-component signals $y=a\sin(2\pi f_1 t+\phi_1)+(1-a)\sin(2\pi f_2 t+\phi_2)$, we also probe the mixing weight `$a$'. Table \ref{tab:parameter-free} shows that a linear probe on $\Hbar$ recovers weight ($a$) with $R^2\!\geq\!0.93$ across trained and pretrained public PFNs.  In particular, the pretrained TabPFN recovers frequency with $R^2=0.993$, and the three further TFMs reach $R^2>0.99$. This shows that spectral organization is a recurring property of the PFNs, and not an artifact of our own training. We further train a higher-capacity MLP probe, which can recover information that is present but nonlinearly entangled.  The last column of Table~\ref{tab:parameter-free} reports the largest absolute gap between the MLP and linear probes over the two tasks. The gap is at most $0.027$ for every task and architecture. On VA-PFN the MLP probe is slightly \emph{worse} than the linear one, consistent with mild over-fitting of the stronger probe. This low gap along with the $R^2 \to 1$ for both probes show that PFNs internally solved the representation-learning problem for the scalar targets we probe and frequency and weights are exposed as approximately affine functions of the coordinates of mean-pooled residual stream $\Hbar$.

\begin{table}[h]
\centering
\setlength{\tabcolsep}{4pt}
\caption{Parameter-free alignment (Sec.~\ref{sec:manifold}) and linear probes for frequency $f$ (single sinusoids) and mixing weight $a$ (two-component signals Sec.~\ref{sec:probing}) on $\Hbar$. $\max|\Delta|$ is the largest $|R^2_{\text{MLP}}-R^2_{\text{linear}}|$ over the two targets. Mean $\pm$ std over 5 seeds is shown; SM-GP: spectral-mixture GP prior. All entries use 250 context points per signal.}
\label{tab:parameter-free}
\begin{tabular}{llccccc}
\toprule
Model & Prior & $\rho_{\Delta}$ & $|r|_{\text{PC}_0}$ & Linear $R^2$ ($f$) & Linear $R^2$ ($a$) & $\max|\Delta|$ \\
\midrule
VA-PFN      & SM-GP   & $0.615 \pm 0.038$ & $0.812 \pm 0.026$ & 0.967 & 0.934 & 0.027 \\
DVA-PFN     & SM-GP   & $0.852 \pm 0.015$ & $0.981 \pm 0.006$ & 0.998 & 0.997 & 0.002 \\
TabPFN 2.5  & tabular & $0.765 \pm 0.021$ & $0.882 \pm 0.010$ & 0.993 & 0.982 & 0.002 \\
TabPFN 3.0  & tabular & $0.788 \pm 0.009$ & $0.926 \pm 0.006$ & 1.000 & 1.000 & $<0.001$ \\
TabICLv2    & tabular & $0.938 \pm 0.008$ & $0.985 \pm 0.002$ & 1.000 & 0.999 & $<0.001$ \\
TabSwift    & tabular & $0.861 \pm 0.005$ & $0.956 \pm 0.004$ & 1.000 & 0.999 & $<0.001$ \\
\bottomrule
\end{tabular}
\end{table}

\textbf{What happens when weights are random?}
To confirm that linear accessibility reflects learned structure rather than trivial input geometry \citep{hewitt2019designing}, we repeat the probing experiment on a randomly initialized (untrained) DVA-PFN as a control. The MLP probe still succeeds ($R^2 \geq 0.99$), but the linear probe collapses to $R^2 = 0.18$ (frequency) and $0.64$ (weight), confirming that training specifically organizes $\Hbar$ into a linearly decodable form \citep{ belinkov2022probing}.

\textbf{Analysis with a non-spectral prior.}
To test whether the organization requires a smooth or spectral prior, we train a nanoTabPFN \citep{pfefferle2025nanotabpfn} from scratch on a decision-tree prior whose targets are piecewise-constant functions of the inputs, with no GP or Fourier component. Over five independently trained models, a linear probe on its frozen $\Hbar$ recovers the frequency of held-out sinusoids with $R^2 = 0.951 \pm 0.022$, against $0.089 \pm 0.025$ for the same five models with random weights. Further, a six-block regression variant of nanoTabPFN reaches $0.989 \pm 0.011$. This is comparable to the PFNs trained on smooth or spectral priors, and the probe $R^2$ rises monotonically with training in every model. We hypothesize that any amortized regressor must infer how fast the target varies with the input, since this sets how far it can borrow strength across context points. Note that TabPFN's effective bandwidth is known to be set by its context \citep{zheng2025tables}, and frequency is the most direct coordinate for such a quantity. 

\textbf{A low-rank structure led by PC$_0$.} On TabPFN at 120 context points, a linear probe on PC$_0$ alone recovers about 80\% of the full probe's $R^2$ on sinusoids (0.714 of 0.889). The probe saturates by 8--16 components, and removing PC$_0$ leaves about a fifth of the $R^2$ (0.196; Table~5, Appendix~B.4). On 5D RBF-GP draws, PC$_0$ alone and the remaining components each carry about half (0.422 and 0.416 of 0.883). The spectral
structure is therefore low-rank and led by PC$_0$. Accessibility weakens for multi-component signals: on DVA-PFN a linear probe for $(f_1,f_2,a_1,a_2)$ reaches $R^2=0.50$, and on TabPFN the amplitudes are not recoverable. Attention pooling recovers part of this (Table~\ref{tab:pooling-ablation}).

\subsection{Non-sinusoidal, Higher-dimensional Inputs}\label{sec:non_sine}
Beyond 1D sinusoids, we repeat the parameter-free probing analysis on functions drawn from GPs with RBF, Mat\'ern-3/2 and Mat\'ern-5/2 kernels in input dimensions $D=1$ to $10$. For each kernel family and dimension we sample 500 functions of 500 points each, with lengthscales $\ell$ log-uniform on $[0.05,\,10]$, pass them through frozen TabPFN, and compute both metrics against the characteristic frequency $f_{\mathrm{char}}\propto 1/\ell$; the proportionality constant does not affect either metric, since both are correlations. Both metrics are positive in every cell: they are weakest at $D=1$--$2$ ($\rho_\Delta$ between 0.33 and 0.66), peak at $D=3$--$5$ (RBF at $D=5$: $\rho_\Delta\!=\!0.82$, $|r|_{\mathrm{PC0}}\!=\!0.89$), and decline gradually, with $|r|_{\mathrm{PC0}}\!\geq\!0.78$ at $D=10$ (Table~\ref{tab:hd_probing}). The decline at high $D$ is consistent with per-axis sampling thinning at a fixed context size; we do not have an explanation for the weaker values at $D=1$--$2$. We note that across depth, $\rho_\Delta$ rises, plateaus and dips slightly in both DVA-PFN and TabPFN (Fig.~\ref{fig:layerwise_manifold}). The spectral structure forms early, within one block in DVA-PFN, and the final blocks appear to shift toward formatting the predictive distribution, consistent with evidence that tabular foundation models form predictions early and refine them later \citep{balef2026onelayer}.

\section{Is the Encoding Causal?}
\label{sec:causal}
As probing is correlational, a high probe $R^2$ shows that information is \emph{present} in a representation, not that the network \emph{uses} it \citep{belinkov2022probing}. We test causality with activation patching \citep{geiger2021causal} and targeted subspace patching on the frozen PFNs, and then repeat the tests without a spectral prior and on real data (Sec.~\ref{sec:real_world}) with more detailed in Appendix~\ref{app:causal}. %

\textbf{Activation Patching:}
\label{sec:patching}
We run two sinusoidal signals $A$ and $B$ (frequency $f_A$ and $f_B$), through the frozen DVA-PFN and cache intermediate representations at every layer. We then \emph{patch} (replace) the representation of signal $A$ with that of signal $B$ at layer $j$ and continue the forward pass. The \textbf{Causal Effect}, $\mathrm{CE} = 1 - \mathrm{MSE}(\hat{y}_{\text{patch}}, \hat{y}_B)/\mathrm{MSE}(\hat{y}_A, \hat{y}_B)$ with $\hat{y}$ the model's predictions, measures how far the prediction moves from $A$ toward $B$; CE $=1$ is a complete transfer. Three intervention sites isolate distinct causal pathways; $\mathbf{h}^{(j)}$ denotes the latent entering block $j$, so $\mathbf{h}^{(1)}=H^{(0)}=\phi_x(X_{\mathrm{q}})$ depends on the $t$-grid alone and $\mathbf{h}^{(j)}=H^{(j-1)}$ for $j>1$. The \textbf{H-patch} ($\mathbf{h}^{(j)}_A \!\leftarrow\! \mathbf{h}^{(j)}_B$) tests our central claim, that the latent $H$ causally carries spectral identity. The \textbf{V-patch} ($\mathbf{v}_A \!\leftarrow\! \mathbf{v}_B$) is a positive control: $V$ encodes only the raw function values in DVA-PFN so replacing it is equivalent to replacing the input data and CE should approach unity. The \textbf{K-patch} ($\mathbf{k}_A \!\leftarrow\! \mathbf{k}_B$) with $K = \phi_x(X_{\mathrm{ctx}})$ is a negative control: $A$ and $B$ share the same $t$-grid in $\sin(2\pi f t + \varphi)$, so $K_A \approx K_B$ and intervention should have negligible effect. We evaluate $n=50$ random pairs ($f \in [0.5, 5.0]$~Hz, minimum gap $\geq 1.0$~Hz) across all six layers in DVA-PFN. We observe three things.

First, replacing $\mathbf{h}^{(j)}$ at any layer $j \geq 2$ shifts the prediction completely to match the donor signal (CE $=0.999$), which establishes that $H$ is a causal carrier of spectral information. Second, for both DVA and TabPFN, spectral encoding emerges in the first attention step (Fig.~\ref{fig:layerwise_manifold}). Patching $\mathbf{h}^{(1)}$ has no effect, since $\mathbf{h}^{(1)}$ encodes the $t$-grid that $A$ and $B$ share. After one block the latent already carries the donor's spectral identity (CE $0.999$ at $L_2$), so a single cross-attention layer suffices to fuse spectral information into $H$, and subsequent layers maintain rather than build this representation. Third, $K$ carries no spectral information at any layer (CE $=0$), corroborating the attention separation in DVA.

\begin{figure}[t]
\centering
\includegraphics[width=0.85\linewidth]{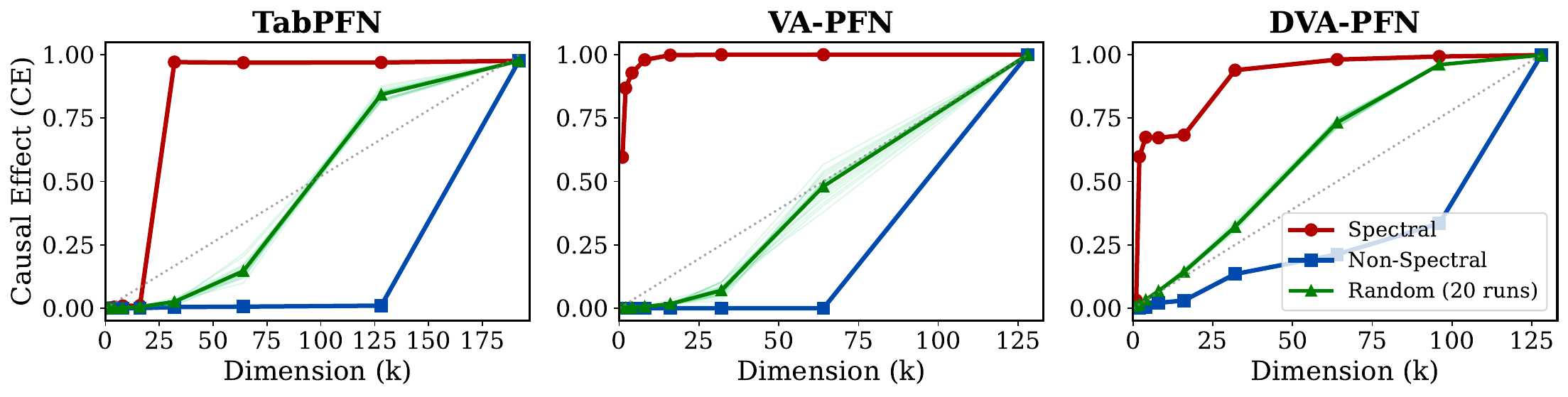}
\vspace{-0.8em}
\caption{Causal dose--response under targeted subspace patching across architectures. Each panel sweeps patched directions $k$ from $1$ to $d$ and reports the causal effect (CE) for spectral (red), non-spectral (blue), or random (green, 20 trials) subspaces.}
\vspace{-1.5em}
\label{fig:combined_patching}
\end{figure}

\textbf{Targeted Subspace Patching:}
\label{sec:subspace}
Full $\Hbar$-patching establishes that spectral information is causally read from the latent, but leaves open a structural question: \textit{is this information distributed diffusely across the $d$-dimensional latent, or concentrated in a separable subspace?} We localize the causally relevant directions and bound their dimensionality with a \textit{dose--response curve} for all three PFNs. We collect $\Hbar$ representations at a fixed block of each PFN for $2{,}500$ signals ($500$ frequencies $\times\,5$ phases) and run PCA on the resulting embeddings. Each principal component is ranked by $|r|$, the absolute Pearson correlation between its projection and the generating frequency: the top-$k$ components define the \emph{spectral subspace}, the bottom-$k$ the \emph{non-spectral subspace}, and $k$ random orthogonal directions a baseline. Unlike distributed alignment search \citep{geiger2024finding}, which learns the subspace by optimization, we choose it by this ranking. We then patch only the selected $k$ PC dimensions of $\Hbar_A$ with the corresponding components of $\Hbar_B$, leaving the remaining $d-k$ dimensions intact, and sweep $k$ from $1$ to $d$ across all three architectures.

Figure~\ref{fig:combined_patching} (left) shows that patching just the top few spectral PCs of TabPFN's $\Hbar$ shifts predictions toward the donor signal, while patching an equal number of non-spectral or random directions changes predictions meaningfully only when $k>100$ out of 192, consistent with the dominant PC$_0$ axis found by probing (Section~3.2). DVA-PFN and VA-PFN replicate the pattern, with CE $\to 1$ much faster than for random and non-spectral baselines.

\subsection{Causality without a Spectral Prior and on Real Data}\label{sec:real_world}
We repeat targeted subspace patching on the tree-prior nanoTabPFN (Sec.~\ref{sec:probing}), using a regression variant of the same prior with five independently trained models per configuration. The same models at initialization serve as untrained controls in Appendix ~\ref{app:noprior_patch}. Principal directions of the last-block latent are ranked by their correlation with frequency, as in previous subsection. In the six-block model, patching the top $16$ of $192$ directions moves the prediction toward the donor with $\mathrm{CE}=0.68\pm0.23$, against $0.06\pm0.04$ for $16$ random directions and $0.01\pm0.01$ for the $16$ least-correlated ones. The three-block model gives $0.67\pm0.22$ for its top $16$ of $96$, against $0.10\pm0.06$ and $0.02\pm0.02$. This ordering holds at every $k$ we sweep (Table~\ref{tab:noprior_patch}). At initialization this intervention is undefined as an untrained model's predictions do not depend on the context, so there are no distinct predictions to move between. We observe that during training, the probe $R^2$ rises monotonically in all 15 models. The gap between top-$16$ and random-$16$ patching is zero through step $100$, rises steeply between steps $250$ and $1{,}000$ as the held-out prior loss falls. 

\textbf{Real datasets:} On real data TabPFN routes the entire effect of its context through a small subspace of its residual stream, and we find this subspace without using any labels. Real datasets have no ground-truth frequency to rank directions by, so here we test the second half of the causal claim, that the context acts through a low-dimensional subspace. We use eight pairs of real datasets with $D$ from 5 to 499  and one control pair whose donor target is the source target permuted over rows (En-perm). The model is run on a source context $A$ and a donor context $B$, and both are queried at the test rows of $A$. At block 12 or block 24 of 24, we fit PCA to sampled 192-dimensional residual-stream vectors, order the directions by explained variance alone, and patch $H \leftarrow H(I-P) + H_B P$, where $P$ projects onto the top $k$ directions. The readout rescales predictions with the target statistics of the source, so even full replacement ($k=192$) reaches a causal effect between 0.58 and 0.94. We therefore report each patch as a fraction of this ceiling. With only 64 of the 192 directions, the top subspace transfers 85 to 95\% of the achievable effect at block 12 in every pair except ISOLET. The same number of random directions transfers 6 to 35\%, and the 64 lowest-variance directions transfer at most 2\% (Table~\ref{tab:real_subspace}). The subspace also stays small as inputs grow. Across the eight real pairs, $k_{50}$, the number of directions needed for half the effect, rises only from 24 to 64 while $D$ grows a hundredfold. This mirrors the frequency-ranked results on synthetic signals (Sec.~\ref{sec:subspace}) and on the tree prior. Because no label enters the ordering, the subspace cannot be selected to fit the outcome, which guards against illusory subspace-patching effects \citep{makelov2024subspace}. On real data this result locates the channel through which the context acts. More details are in Appendix~\ref{app:real}.

\begin{table}[t]
\centering
\small
\setlength{\tabcolsep}{4pt}
\renewcommand{\arraystretch}{0.95}
\caption{Fraction of each pair's full-replacement effect transferred by patching the top 64 variance directions (top) or 64 random directions (rand.) at block 12 or block 24. $k_{50}$ is the smallest $k$ that reaches half of this effect at block 12. Means over 24 draws, with standard error at most 0.05.}
\label{tab:real_subspace}
\vspace{-4pt}
\begin{tabular}{@{}l*{9}{c}@{}}
\toprule
 & Air-5 & ETT-L & ETT-D & Energy & En-perm & Wine & Air-20 & Gas & ISOLET \\
\midrule
$D$              & 5   & 6   & 6   & 8   & 8   & 11  & 20  & 127 & 499 \\
$k_{50}$         & 24  & 24  & 24  & 32  & 16  & 32  & 32  & 48  & 64  \\
\midrule
B$_{12}$ top     & 0.95 & 0.88 & 0.89 & 0.85 & 0.93 & 0.90 & 0.87 & 0.87 & 0.59 \\
B$_{12}$ rand.   & 0.35 & 0.13 & 0.16 & 0.06 & 0.17 & 0.20 & 0.31 & 0.10 & 0.11 \\
B$_{24}$ top     & 0.96 & 0.94 & 0.90 & 0.91 & 0.95 & 0.93 & 0.89 & 0.77 & 0.77 \\
B$_{24}$ rand.   & 0.34 & 0.27 & 0.39 & 0.32 & 0.13 & 0.30 & 0.34 & 0.23 & 0.36 \\
\bottomrule
\end{tabular}
\vspace{-6pt}
\end{table}

\section{Decoding Bayesian Structure: From Latents to Explicit Kernels}
\label{sec:decoder}
If $\Hbar$ encodes the spectral content of the context, an explicit stationary kernel $k(\tau)$, or equivalently its spectral density $S(\omega)$, should be extractable from it. The decoder is not meant to compete with kernel discovery methods, whether iterative \citep{lloyd2014automaticconstructionnaturallanguagedescription,duvenaud2013structurediscoverynonparametricregression} or amortized \citep{bitzer2023amortized}. It is a constructive test that the structure found in Sections~\ref{sec:locate} and~\ref{sec:causal} is rich enough to specify a full covariance function, one of the most demanding read-outs of a representation.

\begin{figure}[t]
\centering
\vspace{-0.7em}
\includegraphics[width=0.95\linewidth]{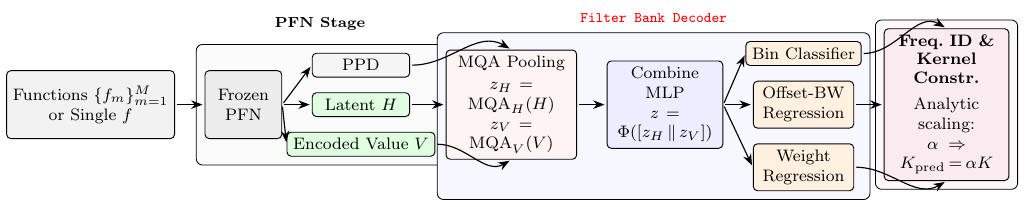}
\vspace{-1em}
\caption{The proposed \textbf{Filter Bank Decoder} (steps described in the text). Multi-query attention pooling helps once spectral complexity exceeds a single component (Table \ref{tab:pooling-ablation} in Appendix). The PFN is never updated; decoder details are in Appendix~\ref{app:decoder}.}
\vspace{-1em}
\label{fig:decoder_architecture}
\end{figure}

\textbf{What is recoverable from data?}
Two facts about identifiability shape our decoder design. \emph{First}, from a single variance-normalized realization from spectral prior, the spectral peak \emph{locations} and \emph{bandwidths} of $S(\omega)$ are recoverable, but the spectral \emph{weights} are identifiable only up to a common multiplicative constant even as $N\!\to\!\infty$. The missing global scale $\alpha$ admits an unbiased plug-in estimator $\hat{\alpha}=\lVert\mathbf{y}\rVert_2^2/\mathrm{tr}(K)$, where $\mathbf{y}$ is the vector of context targets and $K$ is the Gram matrix of the decoded (unscaled) kernel on the context inputs. The covariance used for regression is then $K_{\mathrm{pred}}=\hat{\alpha}K$, so the scale is recovered analytically rather than predicted by the network. \emph{Second}, with $M$ independent realizations from the same prior, the full set $\{w_q,\mu_q,\sigma_q\}$ becomes identifiable. We instantiate a single-realization and a multi-realization decoder, with the PFN frozen throughout (Figure~\ref{fig:decoder_architecture}). Appendix~\ref{app:decoder_all} contains proofs, details and additional results.

\textbf{Decoder pipeline:} The pipeline has six steps. (1) The context passes once through the frozen DVA-PFN, and we collect $H$ at the final block and the value-stream encoding $V$. (2) Two multi-query attention poolers summarize $H$ and $V$ over positions, and an MLP fuses them into one vector. (3) Three heads map this vector to outputs on a fixed grid of frequency bins: an activation probability, a peak offset and bandwidth, and, in the multi-realization decoder, a weight. (4) Bins whose activation probability exceeds a threshold define a spectral mixture $S(\omega)$. (5) Bochner's theorem turns each active bin into a damped cosine, giving $k(\tau)$ in closed form Sec. \ref{sec:bochner}, with analytical global scale $\hat\alpha$. (6) The kernel is used for exact GP regression on the context. The decoder is trained on synthetic draws whose generating spectral densities supply the bin targets. For example, given a 2\,Hz sinusoid, the bin near 2\,Hz activates and the decoded kernel is a slowly damped cosine of period 0.5.

\paragraph{What the decoder reads.}
To ask whether the decoder reads structure from the PFN or infers kernels on its own, we compare it with four variants, scored by kernel reconstruction error (Ker-MSE; mean over 5 seeds). Applying the trained decoder to the latents of a randomly initialized PFN raises kernel error by roughly two orders of magnitude (RBF Ker-MSE $0.519$ vs.\ $0.006$). With the $V$ input removed the decoder matches the full model on every family, and with the $H$ input removed it fails, so the kernel information sits in $H$, not in the value encoding (Sec.~\ref{sec:manifold}). An identical head trained end-to-end on raw $(x, y)$ context matches our decoder, which shows that the frozen latents preserve the information needed to specify the kernel; in our runs, two of three raw-input restarts collapsed to a constant solution while the latent decoder trained stably.

\textbf{Predictive performance with no test-time optimization.}
We evaluate the decoded kernel as a plug-in covariance for GP regression on five kernel families (RBF, Periodic, their product $K_{\text{RP}}$, and spectral mixtures with $Q\!\in\!\{1,4\}$). The decoder receives no information about which kernel family generated each task and it sees only the context $(X, y)$, runs one forward pass through the frozen PFN, and returns an explicit $k(\tau)$. We compare against deep kernel learning (DKL) \citep{wilson2015deepkernellearning}, random Fourier features (RFF) \citep{NIPS2007_013a006f} and Amor-struct \citep{bitzer2023amortized}. The oracle GP uses the true kernel family with hyperparameters refit by marginal likelihood (best of five restarts). Table \ref{tab:cookbook} shows that, despite this fully amortized setting, the decoded kernel has lower MSE than DKL and RFF on Periodic and $K_{\text{RP}}$ and at most about twice their MSE on RBF and the spectral mixtures, while running roughly 200--250$\times$ faster, as those baselines re-optimize per task and our decoder does not. Amor-struct, which receives the kernel family, has lower MSE than the decoder on four of five families and ties on Periodic, at $8\times$ higher latency and with an architecture designed for kernel regression. On negative log predictive density (NLPD), the decoded kernel has the lowest mean of the three learned methods (decoder, DKL, RFF) on all five families, with the largest margins on RBF and the spectral mixtures; the oracle GP is best throughout. On out-of-distribution triangle waves the decoded kernel is comparable to DKL and RFF at 50 and 100 context points.

\begin{table}[t]
\centering\small
\setlength{\tabcolsep}{4pt}
\caption{GP regression MSE on five kernel families with 50 context points ($K_{\text{RP}}$ denotes RBF $\times$ Periodic). The Decoder (single-realization) and PFN columns use DVA-PFN. The decoder serves as a diagnostic extraction, not a general predictive replacement.}
\vspace{-1em}
\label{tab:cookbook}
\begin{tabular}{lcccccc}
\toprule
True kernel & Decoder (Ours) & PFN (DVA-PFN) & Amor-struct. & DKL & RFF & Oracle GP \\
\midrule
RBF                  & $1.1\times10^{-3}$ & $9.7\times10^{-4}$ & $4.6\times10^{-4}$ & $8.4\times10^{-4}$ & $1.0\times10^{-3}$ & $1.7\times10^{-4}$ \\
Periodic             & $1.5\times10^{-3}$ & $1.4\times10^{-3}$ & $1.5\times10^{-3}$ & $6.6\times10^{-3}$ & $6.4\times10^{-3}$ & $8.1\times10^{-4}$ \\
$K_{\text{RP}}$      & $2.7\times10^{-3}$ & $1.1\times10^{-3}$ & $1.6\times10^{-3}$ & $4.9\times10^{-3}$ & $4.6\times10^{-3}$ & $9.5\times10^{-4}$ \\
SM ($Q=1$)           & $1.3\times10^{-3}$ & $4.6\times10^{-4}$ & $4.3\times10^{-4}$ & $6.4\times10^{-4}$ & $8.0\times10^{-4}$ & $2.9\times10^{-4}$ \\
SM ($Q=4$)           & $1.5\times10^{-3}$ & $6.2\times10^{-4}$ & $4.4\times10^{-4}$ & $1.1\times10^{-3}$ & $1.4\times10^{-3}$ & $1.9\times10^{-4}$ \\
\midrule
\textbf{Avg. Time (s)} & \textbf{0.0036} & \textbf{0.0020} & \textbf{0.0288} & \textbf{0.9180} & \textbf{0.7580} & \textbf{3.9460} \\
\bottomrule
\end{tabular}
\vspace{-1em}
\end{table}

Two remaining concerns are 1) whether a PFN is needed at all, and 2) whether the decoded kernel is useful beyond the benchmark. On the first, a classical pipeline can fit a SM to the periodogram or Lomb--Scargle periodogram of the raw context with no PFN. Doing so yields kernels that fit the context set at least as well as ours (lower Ker-MSE) but generalize poorly. GP-MSE on unseen targets is $17$--$78\times$ higher than with the decoded kernel across the four families, because the single-realization weight ambiguity of Theorem~\ref{thm:single-nonident} lets a least-squares fit follow the sampling of one realization. In contrast, the decoder inherits the regularization of the PFN's pretraining prior. On the second concern, the decoded kernel defines a closed-form GP that can be updated without the PFN. A Bayesian optimization loop that calls the PFN once and thereafter runs on the decoded kernel alone reaches a final regret of $0.006\pm0.029$ on four-component SM, against $0.004\pm0.028$ for the oracle GP, at $4.8$\,ms per iteration on CPU; using the PFN itself as the surrogate \citep{muller2022bayesian} costs a GPU forward pass per step. Finally, the same decoder extends beyond the single-realization setting, given $M$ independent realizations, for which Theorem~\ref{thm:multi-ident} makes $S(\omega)$ fully identifiable, decoding error falls with $M$, and decoded kernels stay within $5.5\times$ of the oracle GP's MSE on spectral mixtures in 1D, 5D and 10D. Protocols, tables and figures for all three are in Appendix~\ref{app:classical}--\ref{app:scaling}.

\section{Limitations and Conclusion}
\label{sec:discussion}

\textbf{Limitations:}
Our probes target  the frequency or lengthscale of sinusoids and stationary GP draws. These are the quantities we selected and $\Hbar$ carries other information as well. On real data we locate the channel through which the context acts, not a spectral quantity. Further, the decoder relies on Bochner's theorem and therefore can extracts stationary kernels only. Non-stationary structure would require a different decoder. Also, the decoder is a diagnostic read-out, not a replacement for the PFN and Table~\ref{tab:cookbook} shows a consistent MSE gap, indicating that some predictive information does escapes the spectral-mixture parametrization. We have shown the decoder only on DVA-PFN latents. Our causal account identifies \emph{where} spectral information is stored and \emph{that} it is used, but not \emph{how} attention constructs it.

\textbf{Conclusion.}
\label{sec:conclusion}
We showed that PFN latents encode the spectral content of their context in a linearly decodable structure across seven PFNs, including one trained without any smooth or spectral prior. Also, we showed PFNs (and TFMS) use this structure causally through a low-dimensional subspace. Read out from a frozen DVA-PFN by the Filter Bank Decoder, the structure specifies an explicit stationary kernel that supports GP regression competitive with iterative baselines at a fraction of their cost, and Bayesian optimization on CPU. The early plateau of the structure across depth, our architecture sweep (Appendix~\ref{sec:abl}) and evidence that later layers of TFMs are largely redundant \citep{balef2026onelayer} suggest that distillation and tuning should target the later layers.

\subsection*{Reproducibility statement}
Appendix~\ref{app:models} describes the architectures and pretraining of VA-PFN and DVA-PFN (App.~\ref{app:pfn_pretraining}), the input preprocessing, how $H$, $V$ and $K$ are extracted from each model (App.~\ref{app:representation}), and the full decision-tree prior and training configuration of the nanoTabPFN models (App.~\ref{app:extra_pfns}); the public TFMs are used as released, without fine-tuning. Appendix~\ref{app:probe-setup} specifies the probe signals, both probe architectures with their hyperparameters, and the data splits used in Section~\ref{sec:locate}. Appendix~\ref{app:causal} gives the protocols for activation patching (App.~\ref{app:actpatch}), targeted subspace patching (App.~\ref{app:subspace}, settings in Table~\ref{tab:rbf_hparams}), patching without a spectral prior (App.~\ref{app:noprior_patch}), and the real-data experiments (App.~\ref{app:real}), including the preprocessing of each publicly available dataset and the exact intervention hooks. Complete proofs of Theorems~\ref{thm:single-nonident} and~\ref{thm:multi-ident} and Proposition~\ref{prop:scale-estimator}, with their assumptions, are in Appendix~\ref{app:identifiability}. The decoder architecture, loss, curriculum, training data generation and hyperparameters (Table~\ref{tab:decoder-hyperparams}) are in Appendix~\ref{app:decoder}, and the configurations of all baselines are in Appendix~\ref{app:baselines}. The architecture sweep of Appendix~\ref{sec:abl} lists its full training setup. Experiments use fixed random seeds, and results are reported as mean with standard deviation over seeds, or standard error over draws for the real-data experiments, where stated. 

\bibliography{main}
\bibliographystyle{plainnat}

\newpage
\appendix

\begin{center}
\Large 
    \textbf{Supplementary Material:  Mechanistic Evidence for Spectral Structures in Prior-Data Fitted Networks}
\end{center}

\section{Models and Pretraining}
\label{app:models}
This appendix supports Section~\ref{sec:pfn}: it describes how we pretrain VA-PFN and DVA-PFN, the input preprocessing, and the additional PFNs of Table~\ref{tab:parameter-free}.

\subsection{VA-PFN and DVA-PFN Pretraining}
\label{app:pfn_pretraining}
The DVA and VA PFN is trained on synthetic 1D functions drawn from a stationary GP with a random Spectral Mixture Kernel (SMK). The architecture is a Transformer-based Conditional Neural Process with 6 cross-attention blocks ($4$ heads, $d_{\text{model}} = 128$), trained using categorical cross-entropy over 100 discretized output bins. Our training pipeline is direct extention of the codes given by \citep{muller2022transformers} and \citep{sharma2025dva}

\subsection{Input Preprocessing}
Following the PFN training protocol, we normalize inputs as:
\[
Y_{\text{norm}} = \frac{Y - \mu}{\sigma}, \quad \text{then apply sigmoid activation:}\quad Y_{\text{proc}} = \sigma\left(0.75 \cdot Y_{\text{norm}}\right).
\]
\subsection{Representation Extraction}
\label{app:representation}
We read three tensors from a frozen PFN. $V=\phi_y(Y_{\text{ctx}})$ is the value-stream encoding of the context targets and $K=\phi_x(X_{\text{ctx}})$ the key-stream encoding of the context inputs. $H$ is the residual-stream output of the final attention block at the query positions, as defined in Section~\ref{sec:manifold}; it is the tensor the PFN's prediction head consumes, and the Filter Bank Decoder of Section~\ref{sec:decoder} reads the same $H$ and $V$. For VA-PFN, whose blocks apply self-attention over the concatenated context and query tokens, $H$ is the last block's output restricted to the query tokens. For TabPFN, $H$ is the output tensor of the last transformer block and $V$ the value projection of that block's attention between features; both are mean-pooled over all non-feature axes. For probing, $\Hbar=\tfrac{1}{N}\sum_i H[i]$ is the mean over query positions.

\subsection{Additional PFNs}
\label{app:extra_pfns}

\paragraph{nanoTabPFN on a decision-tree prior.}
We train nanoTabPFN \citep{pfefferle2025nanotabpfn} from scratch on a prior with no GP or Fourier component. Input features are continuous, drawn from normal, uniform, exponential and bimodal mixture distributions. Targets are produced by axis-aligned decision trees, so every target is a piecewise-constant function of the inputs with sharp jumps. The classification variant uses trees of depth 2--5 and 3\% label noise. The regression variant, used for the patching experiment of Section~\ref{sec:real_world}, uses trees of depth 2--4 with continuous leaf values and output noise of 1--8\% of the target standard deviation. We train three configurations, each with five seeds that control both initialization and the prior data stream: a 3-block classifier (width 96, 4 heads), a 3-block regressor of the same size, and a 6-block regressor (width 192, 8 heads), for 10{,}000 steps at batch size 32 with 150 rows per dataset and 100 context rows; checkpoints at initialization serve as untrained controls. As a sanity check, the trained classifiers reach zero-shot in-context accuracy of $100.0 \pm 0.0$ on Iris and $90.7 \pm 0.4$ on Breast Cancer, against $41.6 \pm 3.2$ and $60.2 \pm 10.8$ at initialization, and the regressors reach $R^2 = 0.80$ (3 blocks) and $0.83$ (6 blocks) on held-out draws from the prior. Probing uses the signals and linear ridge probe of Section~\ref{sec:probing} on the frozen mean-pooled latent, fitted on 500 sinusoids and scored on a disjoint set of 500.

\paragraph{Public tabular PFNs.}
TabPFN 2.5 \citep{grinsztajn2025tabpfn}, TabICLv2 \citep{qu2026tabiclv2} and TabSwift \citep{liu2026tabswift} are used as released, frozen and without fine-tuning; each was pretrained on its own synthetic tabular prior. We apply the protocols of Table~\ref{tab:parameter-free} unchanged: the same signals, mean pooling, parameter-free metrics and probes as for TabPFN 2.5, over 5 seeds.

\section{Spectral Organization}
\label{app:spectral}\label{app:mech_interp}\label{app:additional}
This appendix supports Section~\ref{sec:locate}; its subsections follow the order of that section.

\subsection{Probe Inputs and Probes}
\label{app:probe-setup}
For all experiments, we generate sinusoidal signals on $t \in [-1, 1]$ with 250 points.
Frequencies are sampled uniformly from $[0.5, 5.0]$ Hz with random phases
$\phi \sim \mathcal{U}[0, 2\pi]$.
For weighted signals, $$y = a \cdot \sin(2\pi f_1 t + \phi_1) + (1-a)\cdot \sin(2\pi f_2 t + \phi_2).$$

\paragraph{Linear probe.} A single ridge regression ($\alpha=1.0$) with input $\Hbar\in\mathbb{R}^d$ and scalar target. No hidden layers, $d$ parameters. Inputs are standardized (StandardScaler) before fitting. Any $R^2$ this probe achieves is a lower bound on what is linearly encoded in $\Hbar$.

\paragraph{MLP probe.} A three-layer feed-forward network with hidden widths $256\!\to\!128\!\to\!64$. Each layer uses LayerNorm, GELU activation, and dropout ($p=0.1$). Optimized with AdamW (learning rate $10^{-3}$, weight decay $10^{-4}$) using a cosine-annealing schedule (max 500 epochs, batch size 64) and early stopping on validation $R^2$ with patience 50.

\textbf{Data and splits.} Each probing set consists of 2000 synthetic signals (1000 for VA-PFN), each with 250 observation points on $t\in[-1,1]$. Signals are passed through the frozen PFN and $\overline H=\frac1N\sum_i H[i]$ is mean-pooled over positions. Train/validation/test splits are 65\%/15\%/20\%. Context size is 250 points for Table~1 and Figures~4--6, and 120 for the principal-component analyses of Table~5 and Figures~7--9; probe $R^2$ and $|r|_{\mathrm{PC}_0}$ are compared only within a setting.

\subsection{Where Spectral Information Lives}
\label{app:spectral_hv}
\begin{figure}[t]
    \centering
    \begin{subfigure}{\textwidth}
        \centering
        \includegraphics[width=\textwidth]{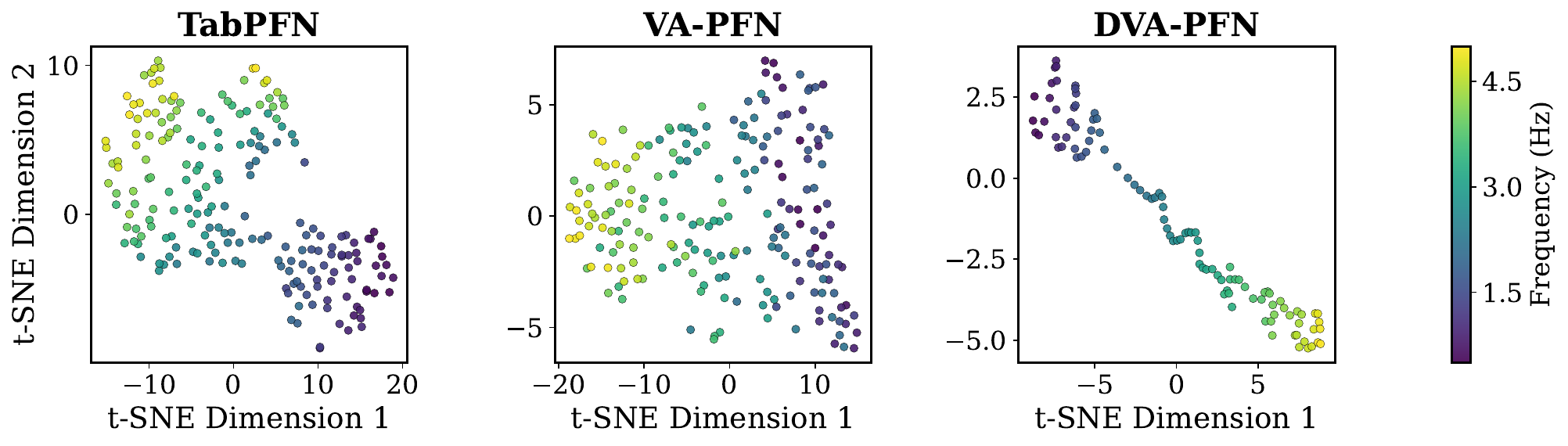}
        \caption{Mean-pooled latent $\Hbar$ manifold.}
        \label{fig:tsne_H}
    \end{subfigure}
    \vspace{0.4cm}
    \begin{subfigure}{\textwidth}
        \centering
        \includegraphics[width=\textwidth]{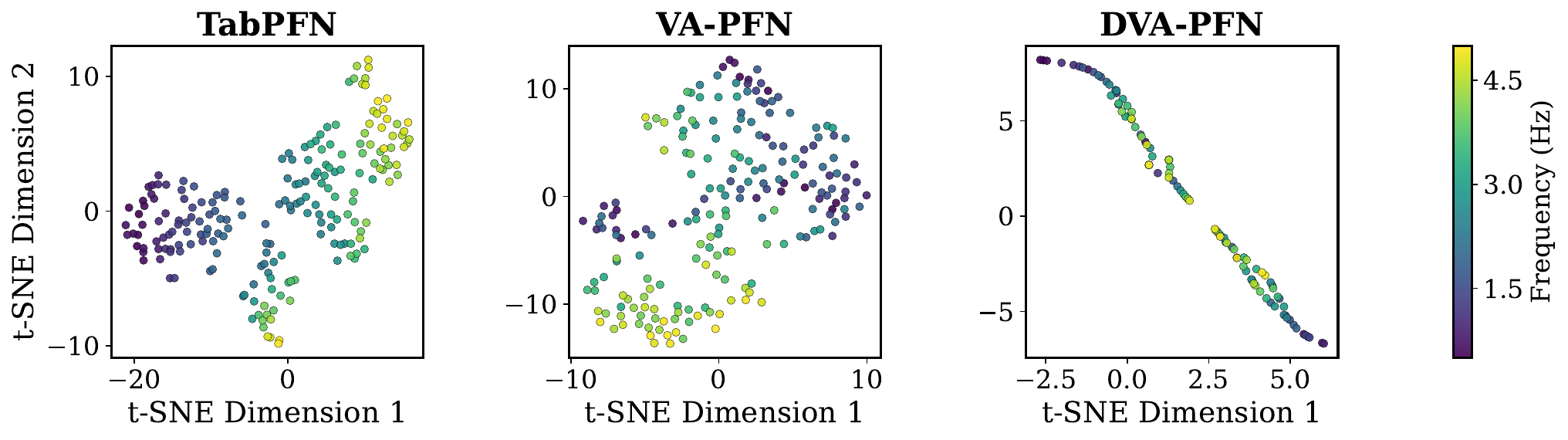}
        \caption{Mean-pooled value encoding $\overline{V}$ manifolds.}
            \label{fig:tsne_V}
    \end{subfigure}
    \caption{Manifold structures colored by frequency obtained via frequecy varying experiment described in Section \ref{sec:locate}. Note the tight spectral clusters in DVA-PFN compared to the smoother manifolds in VA and TabPFN due to diffusion of information inside attention mechanism.}
\end{figure}

Figure~\ref{fig:tsne_H} visualizes the geometry of the mean-pooled $\Hbar$ and $\overline{V}$ in the three core PFNs, and Table~\ref{tab:probing_results_linear} compares linear probes on $H$, $V$ and $H+V$ in DVA-PFN.

\begin{table}[H]
\centering
\caption{Probing R$^2$ scores for spectral parameter extraction using \textbf{linear probes} on DVA-PFN. $H$ consistently dominates $V$ across all targets.}
\label{tab:probing_results_linear}
\begin{tabular}{lccc}
\toprule
Target & $H$ & $V$ & $H+V$ \\
\midrule
Single Frequency & 0.98 & 0.21 & 0.99 \\
Dual Frequencies & 0.96  & 0.00 & 0.96  \\
Full Spectral $(f_1, f_2, a_1, a_2)$ & 0.50 & 0.00 & 0.50 \\
\bottomrule
\end{tabular}
\end{table}

\subsection{Linear and Nonlinear Probes}
\label{app:spectral_probes}
\begin{figure}[H]
    \centering
\includegraphics[width=\textwidth]{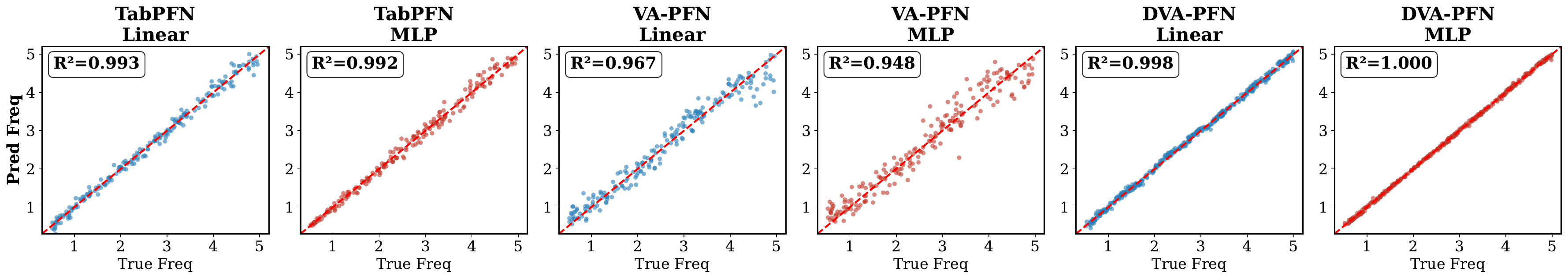}
    \caption{Frequency Probing: Linear and MLP Relational Scatter Plots across PFN architectures.}
    \label{fig:probing_freq_linear}
\end{figure}

\begin{figure}[H]
    \centering
\includegraphics[width=\textwidth]{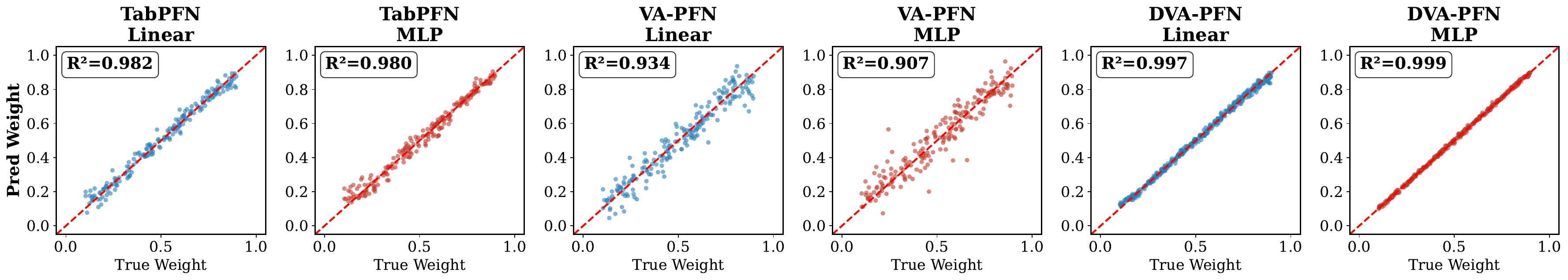}
    \caption{Weight Probing: Linear and MLP Relational Scatter Plots across PFN architectures.}
    \label{fig:probing_weight_linear}
\end{figure}

\begin{figure}[H]
  \centering
  \includegraphics[width=0.75\linewidth]{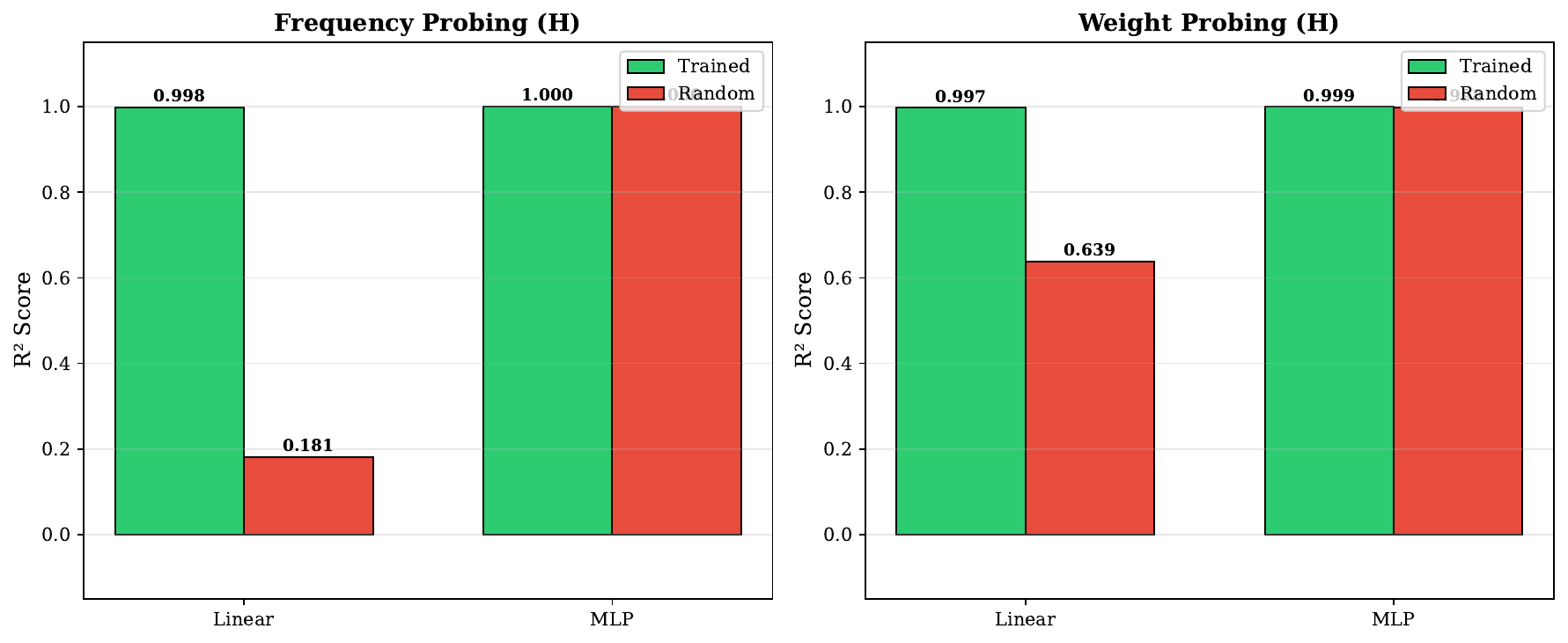}
  \caption{Control experiment: probing on trained vs.\ randomly initialized DVA-PFN. Linear and MLP probes are trained to recover frequency (left) and mixing weight (right) from frozen $\Hbar$. On the trained network both probes succeed ($R^2 \geq 0.99$). On the randomly initialized network, the MLP probe still recovers both targets---consistent with its capacity to learn the mapping itself~\citep{hewitt2019designing}---but the linear probe fails ($R^2 = 0.18$ for frequency, $0.64$ for weight), confirming that linear accessibility is a consequence of learned representational structure, not of trivial input geometry.}
  \label{fig:control}
\end{figure}

Figures~\ref{fig:probing_freq_linear} and~\ref{fig:probing_weight_linear} show linear and MLP probe predictions for frequency and for the mixing weight of two-component signals in the three core PFNs, and Figure~\ref{fig:control} compares both probes on a trained and a randomly initialized DVA-PFN.

\subsection{A Low-Rank Structure Led by PC$_0$}
\label{app:pc_ablation}
\begin{figure}[H]
    \centering
    \includegraphics[width=\linewidth]{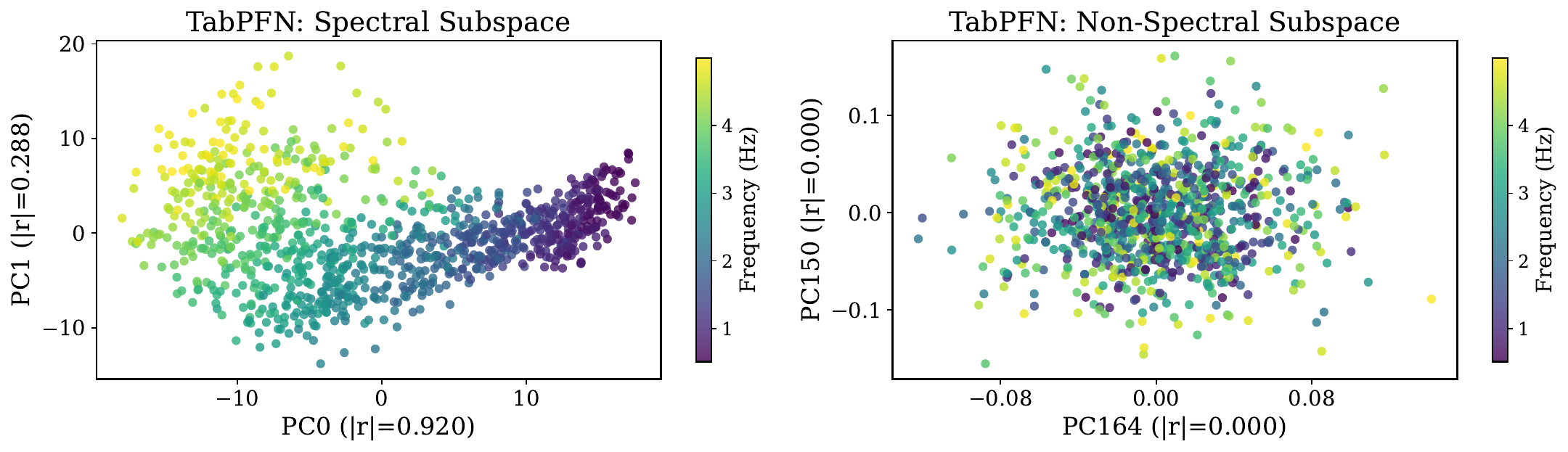}
    \caption{PCA of TabPFN embeddings}
    \label{fig:tabpfn_pca}
\end{figure}

\begin{figure}[H]
    \centering
    \includegraphics[width=\linewidth]{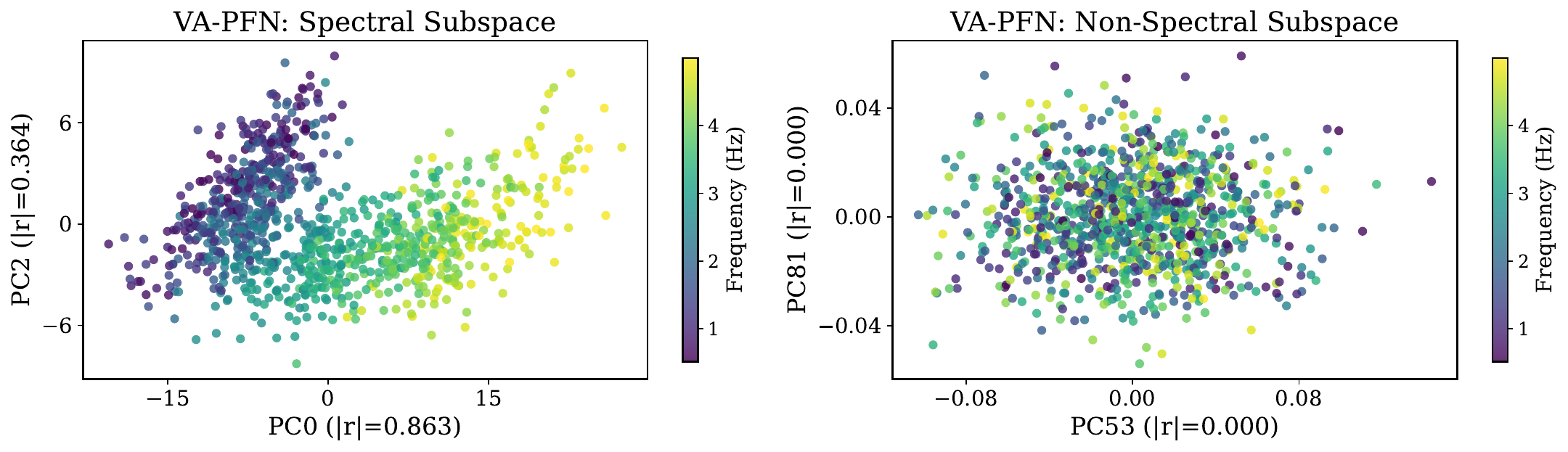}
    \caption{PCA of VA-PFN embeddings}
    \label{fig:vabpfn_pca}
\end{figure}

\begin{figure}[t]
    \centering
    \includegraphics[width=\linewidth]{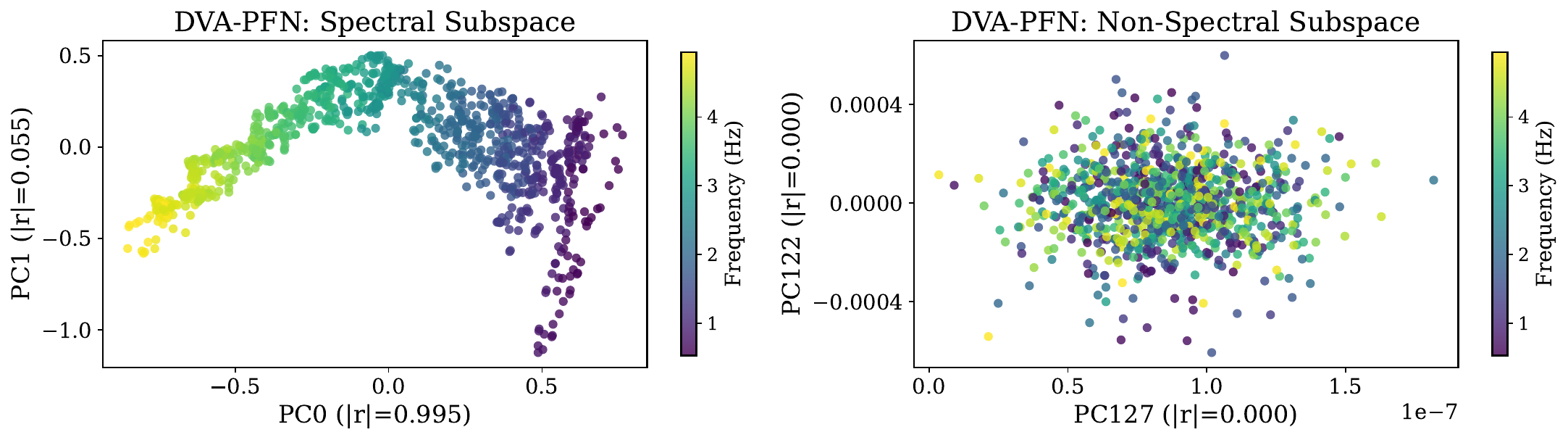}
    \caption{PCA of DVA-PFN embeddings}
    \label{fig:dvapfn_pca}
\end{figure}

Table~\ref{tab:pc_ablation} probes the characteristic frequency from the top-$k$ principal components of TabPFN's mean-pooled $\Hbar$ ($d=192$), and from the same components with PC$_0$ removed. Linear ridge probes use the PCA scores as inputs, with PCA fitted on the training split; $R^2$ is the mean over 5 seeds.
Figures~\ref{fig:tabpfn_pca}--\ref{fig:dvapfn_pca} plot each PFN's embeddings on its most and least frequency-correlated principal components.

\begin{table}[h]
\centering\small
\caption{Probe $R^2$ from the top-$k$ principal components of $\overline H$ on TabPFN, with and without PC$_0$, at 120 context points per signal (Table~1 uses 250).}
\label{tab:pc_ablation}
\begin{tabular}{rcccc}
\toprule
 & \multicolumn{2}{c}{1D sinusoids} & \multicolumn{2}{c}{5D RBF-GP} \\
\cmidrule(lr){2-3}\cmidrule(lr){4-5}
$k$ & top-$k$ & PC$_0$ removed & top-$k$ & PC$_0$ removed \\
\midrule
1 (PC$_0$ alone) & 0.714 & ---   & 0.422 & ---   \\
4                & 0.868 & 0.170 & 0.863 & 0.402 \\
8                & 0.869 & 0.171 & 0.875 & 0.406 \\
16               & 0.879 & 0.184 & 0.880 & 0.413 \\
64               & 0.889 & 0.196 & 0.883 & 0.416 \\
192 (full)       & 0.889 & 0.196 & 0.883 & 0.416 \\
\bottomrule
\end{tabular}
\end{table}


\subsection{Multi-Component Signals and Pooling}
\label{app:spectral_multi}
\begin{figure}[H]
    \centering
    \includegraphics[width=\linewidth]{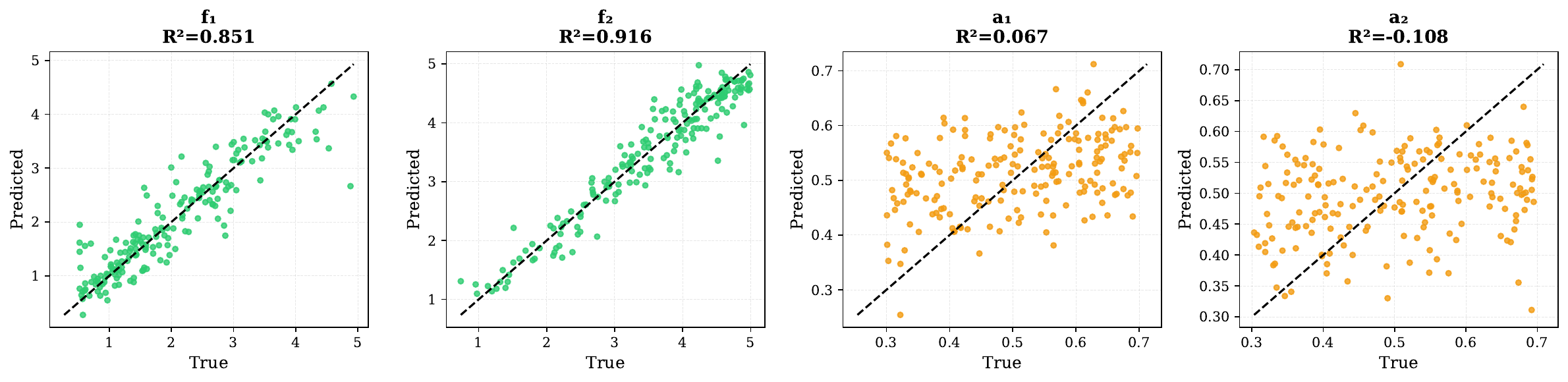}
    \caption{Probing performance on multi-component signals shows successful frequency extraction ($f_1, f_2$) but poor relative amplitude prediction ($a_1, a_2$) on TabPFN.}
    \label{fig:tabpfn_scatter_multi}
\end{figure}

Figure~\ref{fig:tabpfn_scatter_multi} shows TabPFN's linear probes for two-component signals, and Table~\ref{tab:pooling-ablation} compares mean and attention pooling on DVA-PFN.

\begin{table}[H]
\centering\small
\caption{Pooling ablation on $\Hbar$ (MLP probe, $R^2$) on DVA-PFN. Mean pooling degrades sharply as spectral complexity grows, while multi-query attention pooling recovers an increasing fraction of the lost signal. $V$-only probes yield $R^2\!=\!0$ at every difficulty and are omitted. $\Delta$ reports the gain of attention pooling over mean pooling on $H\!+\!V$.}
\label{tab:pooling-ablation}
\begin{tabular}{lccccc}
\toprule
Task & Mean($H$) & Mean($H\!+\!V$) & Attn($H$) & Attn($H\!+\!V$) & $\Delta$ \\
\midrule
Easy\ \ (1 param)      & 0.999 & 1.000 & 1.000 & 1.000 & $+0.000$ \\
Medium (2 params)      & 0.982 & 0.983 & 0.991 & 0.991 & $+0.008$ \\
Hard\ \ (4 params)     & 0.560 & 0.564 & 0.610 & 0.602 & $+0.038$ \\
Very Hard (6 params)   & 0.333 & 0.342 & 0.404 & 0.399 & $+0.057$ \\
\bottomrule
\end{tabular}
\end{table}

\subsection{GP Draws in One to Ten Dimensions}
\label{app:spectral_hd}
\begin{figure}[H]
\centering
\includegraphics[width=\textwidth]{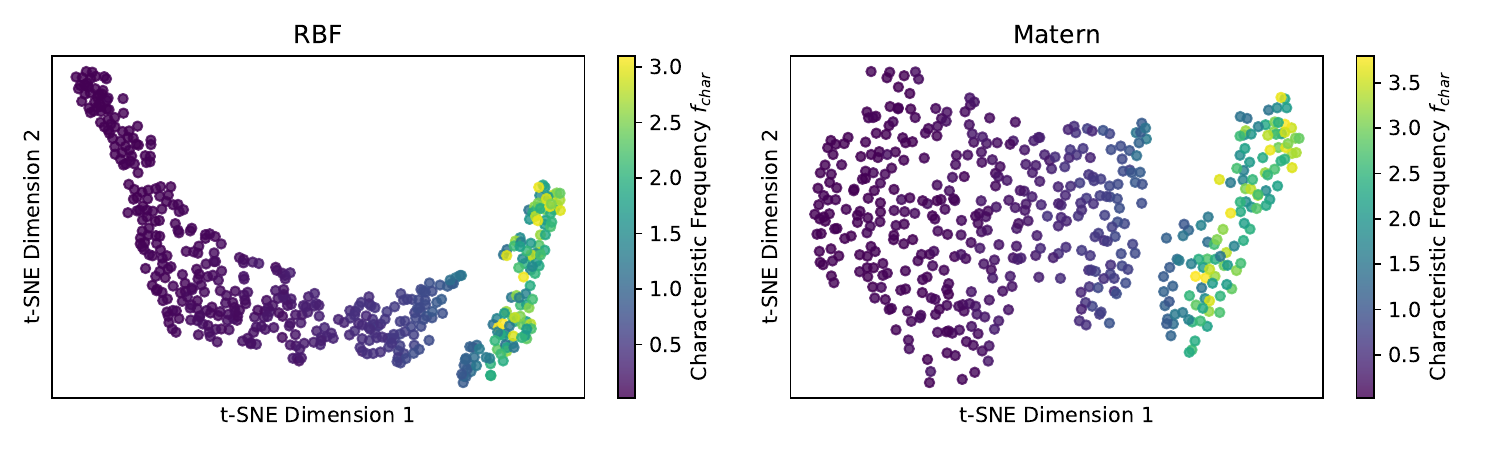}
\caption{t-SNE projections of mean-pooled TabPFN embeddings $\Hbar$ for 500 functions drawn from 5D GPs with RBF (left) and Mat\'ern-3/2 (right) kernels, colored by characteristic frequency $f_{\mathrm{char}}=1/(2\pi\ell)$. Despite the absence of any sinusoidal structure in the generating process, embeddings organize smoothly by spectral scale, consistent with the quantitative metrics in Table~\ref{tab:hd_probing}.}
\label{fig:hd_tsne}
\end{figure}

Table~\ref{tab:hd_probing} gives both parameter-free metrics for every kernel family and input dimension, and Figure~\ref{fig:hd_tsne} shows the embeddings at $D=5$.

\begin{table}[h]
\centering
\small
\caption{Parameter-free probing metrics for GP functions on frozen TabPFN across input dimensions $D$ (500 functions of 500 points per cell, $\ell$ log-uniform on $[0.05, 10]$). At $D=5$ the values agree with an earlier independent run (RBF 0.815 / 0.900; Mat\'ern-3/2 0.861 / 0.928).}
\label{tab:hd_probing}
\begin{tabular}{rcccccc}
\toprule
 & \multicolumn{2}{c}{RBF} & \multicolumn{2}{c}{Mat\'ern-3/2} & \multicolumn{2}{c}{Mat\'ern-5/2} \\
\cmidrule(lr){2-3} \cmidrule(lr){4-5} \cmidrule(lr){6-7}
$D$ & $\rho_\Delta$ & $|r|_{\mathrm{PC0}}$ & $\rho_\Delta$ & $|r|_{\mathrm{PC0}}$ & $\rho_\Delta$ & $|r|_{\mathrm{PC0}}$ \\
\midrule
1  & 0.42 & 0.71 & 0.33 & 0.67 & 0.48 & 0.75 \\
2  & 0.66 & 0.79 & 0.48 & 0.68 & 0.55 & 0.72 \\
3  & 0.81 & 0.93 & 0.83 & 0.90 & 0.78 & 0.91 \\
4  & 0.81 & 0.90 & 0.88 & 0.95 & 0.82 & 0.92 \\
5  & 0.82 & 0.89 & 0.88 & 0.94 & 0.83 & 0.91 \\
6  & 0.77 & 0.88 & 0.83 & 0.91 & 0.78 & 0.88 \\
7  & 0.70 & 0.84 & 0.78 & 0.89 & 0.72 & 0.86 \\
8  & 0.64 & 0.82 & 0.73 & 0.86 & 0.66 & 0.82 \\
9  & 0.59 & 0.80 & 0.69 & 0.83 & 0.62 & 0.81 \\
10 & 0.55 & 0.78 & 0.65 & 0.82 & 0.57 & 0.79 \\
\bottomrule
\end{tabular}
\end{table}

\subsection{Depth Profile}
\label{app:spectral_depth}
\begin{figure}[H]
\centering
\includegraphics[width=0.75\textwidth]{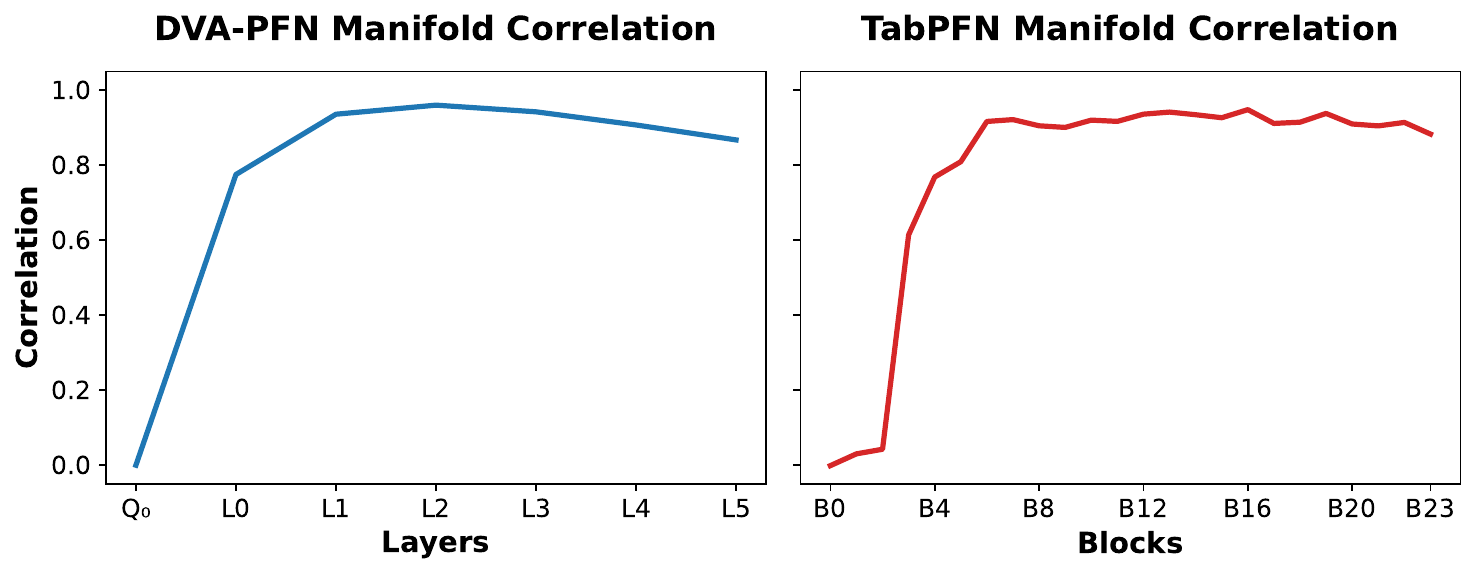}
\caption{Layer-wise correlation $\rho_\Delta$ between embedding distances and generating-frequency distances. DVA-PFN (left) peaks at L2 ($\rho_\Delta=0.95$) and TabPFN (right) plateaus near $0.93$ by B12. Both architectures show a mild decline in the final layers, consistent with a shift toward posterior formatting.}
\label{fig:layerwise_manifold}
\end{figure}

Figure~\ref{fig:layerwise_manifold} tracks $\rho_\Delta$ across the layers of DVA-PFN and TabPFN.

\section{Causal Interventions}
\label{app:causal}\label{app:interpretability_results}
This appendix supports Section~\ref{sec:causal}; its subsections follow the order of that section.

\subsection{Activation Patching}
\label{app:actpatch}
Table~\ref{tab:exp_f} gives the layer-wise causal effect for the sinusoid pairs of Section~\ref{sec:patching}. We repeat the experiment on functions drawn from an RBF-kernel GP.

\begin{table}[h]
\centering
\caption{Layer-wise Causal Effect (CE) for activation patching. Mean $\pm$ s.d.\ over $n=50$ pairs; $p$-values from paired $t$-tests (H vs.\ K).}
\label{tab:exp_f}
\small
\begin{tabular}{@{}lcccc@{}}
\toprule
\textbf{Layer} & \textbf{H-patch CE} & \textbf{V-patch CE} & \textbf{K-patch CE} & \textbf{$p$ (H vs.\ K)} \\
\midrule
L1 & $0.000 \pm 0.000$ & $0.999 \pm 0.002$ & $0.000 \pm 0.000$ & n.a. \\
L2 & $0.999 \pm 0.002$ & $0.999 \pm 0.002$ & $0.000 \pm 0.000$ & $< 10^{-133}$ \\
L3--L6 & $0.999 \pm 0.002$ & $0.999 \pm 0.002$ & $0.000 \pm 0.000$ & $< 10^{-133}$ \\
\bottomrule
\end{tabular}
\end{table}

\label{app:rbf_patching}

\label{app:rbf_probing}

\begin{figure}[h]
  \centering
  \includegraphics[width=\linewidth]{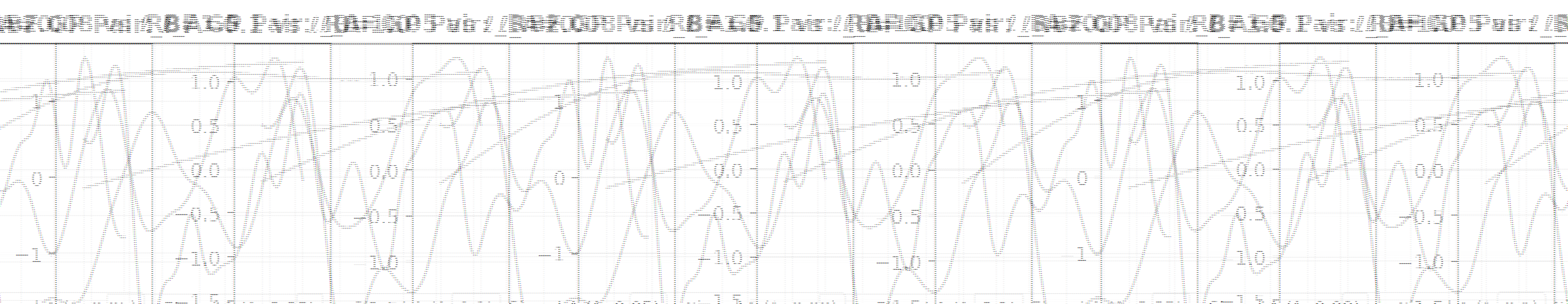}
  \caption{%
    Representative RBF-GP signal pairs used for activation patching.
    Small $\ell$ (red) produces high-frequency wiggly functions;
    large $\ell$ (blue) produces smooth slowly-varying functions.
    The large visual separation ensures a strong baseline
    $\mathrm{MSE}(\hat{y}_A, \hat{y}_B)$, making the causal effect
    measurement well-conditioned.%
  }
  \label{fig:rbf_signals}
\end{figure}

Functions are drawn from a zero-mean GP with RBF kernel
$k(x,x') = \exp\!\bigl(-\tfrac{1}{2}(x-x')^2/\ell^2\bigr)$ on a fixed
grid of $N{=}200$ equally-spaced points in $[-1,1]$.
Lengthscales are sampled log-uniformly: $\ell\sim\mathrm{LogUniform}(0.05,\,2.0)$.
All realisations use fixed random seeds, making the dataset fully deterministic
and reproducible.
Figure~\ref{fig:rbf_signals} shows representative pairs at the extremes of the
$\ell$ range used in the activation patching experiment.

Table~\ref{tab:rbf_act_patch} reports layer-wise causal effect (CE) for
DVA-PFN under H-, V-, and K-patching, averaged over three signal pairs.
CE is defined as
\begin{equation}\label{eq:ce}
  \mathrm{CE} = 1 - \frac{\mathrm{MSE}(\hat{y}_{\mathrm{patched}},\,\hat{y}_B)}
                          {\mathrm{MSE}(\hat{y}_A,\,\hat{y}_B)},
\end{equation}
where $\hat{y}_A,\hat{y}_B$ are the unpatched model predictions
for contexts $y_A$ and $y_B$ respectively (using model predictions rather
than raw GP realisations removes sample-noise from the denominator).

\begin{table}[h]
\centering
\caption{Layer-wise CE for DVA-PFN on RBF-GP pairs
         ($\ell_A\!=\!0.05,\,\ell_B\!=\!2.0$).
         H-patch reaches CE$\approx$1 by layer~2; K-patch gives exactly 0.}
\label{tab:rbf_act_patch}
\small
\begin{tabular}{lccccccc}
\toprule
Patch type & L1 & L2 & L3 & L4 & L5 & L6 \\
\midrule
H (causal)       & 0.00 & \textbf{1.00} & 1.00 & 1.00 & 1.00 & 1.00 \\
V (+control)     & 1.00 & 1.00 & 1.00 & 1.00 & 1.00 & 1.00 \\
K ($-$control)   & 0.00 & 0.00 & 0.00 & 0.00 & 0.00 & 0.00 \\
\bottomrule
\end{tabular}
\end{table}

\noindent

\subsection{Targeted Subspace Patching}
\label{app:subspace}
Table~\ref{tab:exp_g} gives the DVA-PFN values behind Figure~\ref{fig:combined_patching}; for that sweep the intervention block is cross-attention block 2 of 6 for DVA-PFN, transformer block 12 of 24 for TabPFN, and the last of the 6 self-attention blocks for VA-PFN. For the RBF-GP functions of Appendix~\ref{app:actpatch}, Figure~\ref{fig:subspace_rbf} gives the dose-response curves and Table~\ref{tab:rbf_hparams} the settings.

\begin{table}[t]
\centering
\caption{Targeted subspace patching on DVA-PFN: causal effect as a function of subspace dimensionality $k$ (out of $d=128$). Selectivity $=$ Spectral CE / Random CE. }
\label{tab:exp_g}
\small
\begin{tabular}{@{}rccccc@{}}
\toprule
$k$ & \textbf{\% of dims} & \textbf{Spectral CE} & \textbf{Non-spectral CE} & \textbf{Random CE} & \textbf{Selectivity} \\
\midrule
1   & 0.78\% & $0.026$          & $0.000$ & $0.007$ & --- \\
2   & 1.56\% & $\mathbf{0.592}$ & $0.008$  & $0.018$ & $\mathbf{33.9\times}$ \\
4   & 3.13\% & $\mathbf{0.676}$ & $0.017$  & $0.032$ & $21.3\times$ \\
8   & 6.25\% & $0.676$          & $0.023$  & $0.066$ & $10.3\times$ \\
16  & 12.5\% & $0.689$          & $0.115$  & $0.149$ & $4.6\times$ \\
32  & 25.0\% & $\mathbf{0.977}$ & $0.135$  & $0.326$ & $3.0\times$ \\
128 & 100\%  & $0.999$          & $0.999$  & $0.999$ & $1.0\times$ \\
\bottomrule
\end{tabular}
\end{table}

Prior to the dose-response experiment, we verify that PCA identifies a meaningful spectral subspace.

\begin{itemize}[leftmargin=1.5em,itemsep=2pt]
  \item \textbf{DVA-PFN} ($d{=}128$): top-5 PC correlations with $\ell$
    are $[0.30,\,0.28,\,0.26,\,0.19,\,0.18]$.
    Information is distributed across many PCs—consistent with DVA-PFN's
    narrower training distribution—yet concentrated enough that the top-64
    PCs account for all causal transfer (CE$_{\mathrm{spec},\,k=64}=0.93$).

  \item \textbf{TabPFN} ($d{=}192$): PC$_0$ alone achieves $|r|{=}0.882$ and explains $73.6\%$ of embedding variance. The remaining PCs fall off rapidly ($|r|\le 0.21$), indicating a strongly
    dominant single axis of structural variation.
\end{itemize}

\begin{table}[h]
\centering
\caption{Hyperparameters for non-sinusoidal patching experiments.}
\label{tab:rbf_hparams}
\small
\begin{tabular}{ll}
\toprule
Parameter & Value \\
\midrule
Input grid size $N$            & 200 \\
Lengthscale range $[\ell_{\min},\ell_{\max}]$ & $[0.05,\,2.0]$ \\
Probe set size                 & 300 signals (log-uniform $\ell$) \\
Patch pairs                    & 30 (min gap $|\ell_A-\ell_B|\ge 0.8$) \\
Subspace dims $k$              & $\{1,2,4,8,16,32,64,128\}$ \\
DVA-PFN intervention layer     & Cross-attention block 2 (of 6) \\
TabPFN intervention layer      & Transformer block 24 (final; blocks 1--24) \\
Random seed                    & Fixed (all experiments deterministic) \\
\bottomrule
\end{tabular}
\end{table}

\begin{figure}[t]
  \centering
  \includegraphics[width=\linewidth]{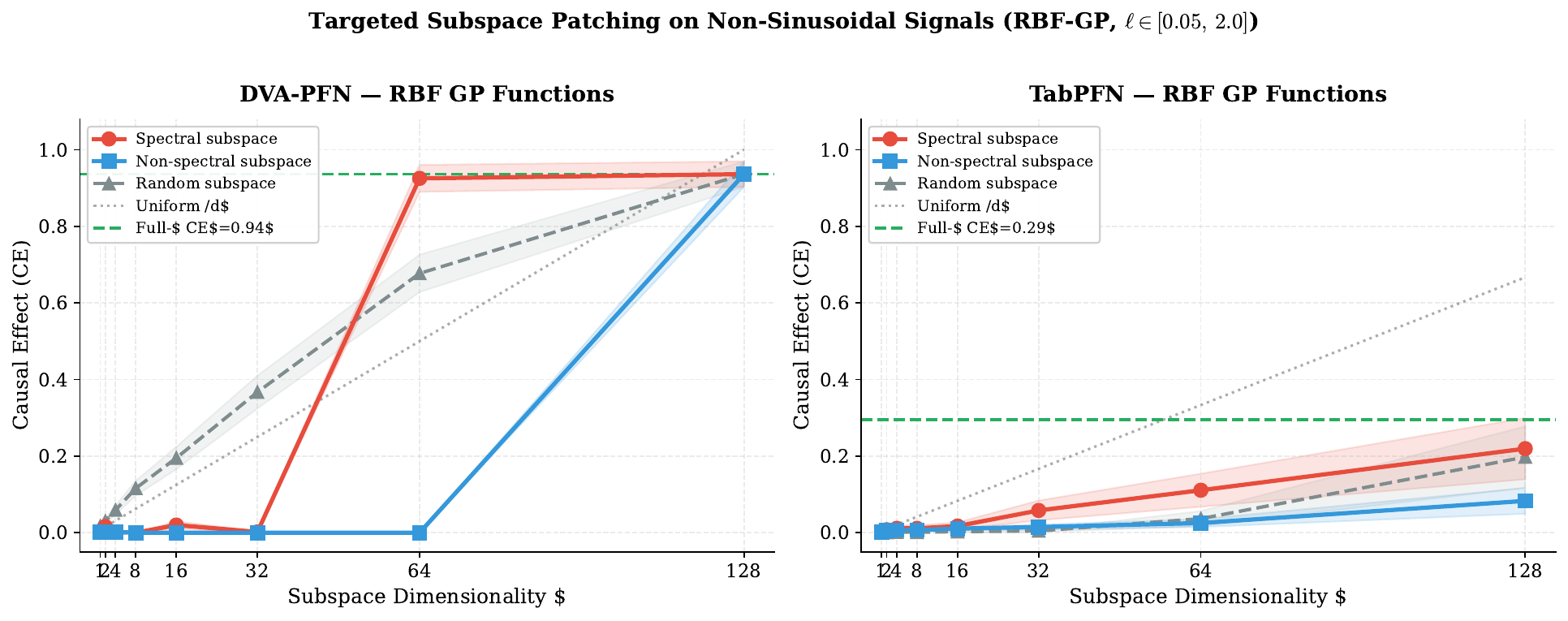}
  \caption{%
    \textbf{Targeted subspace patching on RBF-GP functions.}
    Causal Effect (CE) as a function of patching dimensionality~$k$ for DVA-PFN (\emph{left}) and TabPFN (\emph{right}).
    \textcolor[HTML]{e74c3c}{\textbf{Red}}: spectral subspace
    (PCs most correlated with $\ell$).
    \textcolor[HTML]{3498db}{\textbf{Blue}}: non-spectral subspace (bottom-$k$ PCs).
    \textbf{Grey} triangles: random $k$-dimensional subspace.
    Dashed grey: uniform $k/d$ baseline.
    Dashed green: full-$H$ replacement ceiling.
    Shaded bands: 95\% confidence intervals over 30 signal pairs.
    The spectral subspace dominates at every $k$ for both architectures,
    demonstrating that causal structural information is compactly organised
    beyond the sinusoidal training distribution.%
  }
  \label{fig:subspace_rbf}
\end{figure}

\subsection{Patching without a Spectral Prior}
\label{app:noprior_patch}
The tree-prior nanoTabPFN models are described in Appendix~\ref{app:extra_pfns}. Probe signals are the 500 sinusoids of Section~\ref{sec:probing} with 100 context points. Subspace patching follows Section~\ref{sec:subspace}: PCA is fitted on the mean-pooled last-block latent of the 500 signals, directions are ranked by $|r|$ with frequency, and for each donor--target pair the top-$k$, bottom-$k$ or $k$ random directions of the target's last-block residual stream are replaced by the donor's, at every position, before the prediction head; 30 pairs per model. CE follows Eq.~\eqref{eq:ce} and is reported only over pairs whose unpatched predictions differ by more than $0.05$ per point in root mean square, since CE is undefined when $\hat{y}_A\approx\hat{y}_B$; the fraction of usable pairs is $0.91$--$0.97$ for trained models and $0.007$--$0.02$ at initialization, where predictions are nearly constant. Table~\ref{tab:noprior_patch} gives the dose--response at 10{,}000 steps.

\begin{table}[h]
\centering\small
\caption{Targeted subspace patching of tree-prior nanoTabPFN at the last block after 10{,}000 steps: CE for the top-$k$, $k$ random and bottom-$k$ frequency-ranked directions, mean $\pm$ std over five trained models. $d=96$ for the 3-block model and $192$ for the 6-block model.}
\label{tab:noprior_patch}
\begin{tabular}{rccc|ccc}
\toprule
 & \multicolumn{3}{c|}{3-block regressor} & \multicolumn{3}{c}{6-block regressor} \\
$k$ & top & random & bottom & top & random & bottom \\
\midrule
1  & $0.14\pm0.11$ & $0.11\pm0.07$ & $0.00\pm0.00$ & $0.04\pm0.02$ & $0.02\pm0.02$ & $0.00\pm0.00$ \\
4  & $0.34\pm0.23$ & $0.01\pm0.00$ & $0.01\pm0.01$ & $0.14\pm0.08$ & $0.00\pm0.00$ & $0.01\pm0.01$ \\
16 & $0.67\pm0.22$ & $0.10\pm0.06$ & $0.02\pm0.02$ & $0.68\pm0.23$ & $0.06\pm0.04$ & $0.01\pm0.01$ \\
64 & $0.94\pm0.02$ & $0.64\pm0.18$ & $0.07\pm0.03$ & $0.88\pm0.03$ & $0.09\pm0.02$ & $0.05\pm0.03$ \\
\bottomrule
\end{tabular}
\end{table}

\subsection{Real data}
\label{app:real}

\paragraph{Model.} We use the package-default \texttt{TabPFNRegressor} v2.5:
24 transformer blocks, model width 192, and an 8-member ensemble. We intervene
at block 12 and at block 24, the final block, counting blocks from 1.

\paragraph{Dataset pairs.} Each pair gives a source and a donor dataset with
the same input dimension $D$. Airline Passengers and Milk Production are
first-differenced, $z$-normalized and lag-embedded with 5 or 20 lags
($D=5, 20$). ETTh1 and ETTh2 are used both in levels and first-differenced
($D=6$). Energy efficiency pairs the heating-load target with the cooling-load
target ($D=8$), and, as a control, the heating-load target with a random
permutation of itself over rows, so that the donor's target carries no
relation to its inputs. We further use Wine red vs.\ white ($D=11$),
gas-sensor drift batch 1 vs.\ batch 10 ($D=127$), and ISOLET with a
speaker-disjoint split ($D=499$). Each draw samples 100 context rows and 30
test rows.

\paragraph{Intervention.} TabPFN's block output is a tensor over rows,
feature groups and width; we call each 192-dimensional vector in it a cell. For
a source context $A$ and a donor context $B$, the model fitted on each context
predicts at $A$'s test rows, so that $\hat y_A$, $\hat y_B$ and the patched
prediction are defined at the same inputs. With $V_k \in \mathbb{R}^{k \times
192}$ holding $k$ selected principal directions and $P = V_k^\top V_k$, the
hook applies
\begin{equation}
H \leftarrow H\,(I - P) + H_B\,P
\end{equation}
to every cell of the block output. Ensemble members differ in their
feature-group shapes, so donor tensors are matched to each member, and we
verify that the hook fires once per member on every forward pass.

\paragraph{Subspace.} Principal directions are computed from the cell vectors
themselves, the objects the patch replaces, with 120 cells sampled per context
from $N$ contexts of each dataset in the pair, and ordered by explained
variance. No target or dataset label enters the ordering. We compare the top-$k$
directions with the bottom-$k$ and with $k$ random orthonormal directions; a
random $k$-dimensional subspace captures $k/192$ of the variance in
expectation.

\paragraph{Causal effect.} For each draw we compute
$\mathrm{CE}_k = 1 - \lVert \hat y_{\mathrm{patch}} - \hat y_B \rVert /
\lVert \hat y_A - \hat y_B \rVert$, which is stricter than
Eq.~(\ref{eq:ce}), since $1-r \le 1-r^2$ for $r \in [0,1]$. Draws are kept
only when $\lVert \hat y_A - \hat y_B \rVert$ exceeds 0.1 times its median
over draws and 0.05 per point in root mean square, and CE is clipped to
$[0,1]$. Replacing the entire block output does not reproduce $\hat y_B$,
because the readout de-standardizes predictions with the source context's
target statistics. The full-replacement effect $\mathrm{CE}_{\mathrm{full}}$
is therefore 0.58 to 0.94 depending on the pair, and we report
$\mathrm{CE}_k / \mathrm{CE}_{\mathrm{full}}$, the fraction of the achievable
effect.

\paragraph{Draws.} The points $k = 32$ and $k = 64$ use 24 donor--target
draws per pair and three random subspaces. The full sweep
$k \in \{1, 4, 8, 16, 24, 32, 48, 64, 96, 192\}$ uses 6 draws per pair (12
for Wine) and reproduces the 24-draw values to within a few points.

\paragraph{Results.} Table~\ref{tab:real_k50} gives the smallest $k$
reaching 50\% and 80\% of the full effect at both blocks, and
Tables~\ref{tab:real_b12} and~\ref{tab:real_b24} give the 24-draw results.
At $k = 64$ the top-variance subspace carries 59--95\% of the full effect at
block 12 and 77--96\% at block 24, against 6--39\% for a random subspace of
the same size and at most 2\% for the lowest-variance one. At block 12 the
top subspace exceeds the random one by a factor of 2.7 to 15. Every
top-versus-random gap exceeds its standard error many times over. The
permuted-target pair concentrates as strongly as any other, so the subspace
carries the context's influence whether or not the donor's target is related
to its inputs.

\begin{table}[h]
\centering
\small
\caption{Smallest $k$ (of 192) whose top-variance patch reaches 50\% ($k_{50}$)
or 80\% ($k_{80}$) of the full-replacement effect, read from the sweep grid,
and the full-replacement effect $\mathrm{CE}_{\mathrm{full}}$ of each pair.}
\label{tab:real_k50}
\begin{tabular}{lrccccc}
\toprule
 & & \multicolumn{2}{c}{Block 12} & \multicolumn{2}{c}{Block 24} & \\
\cmidrule(lr){3-4} \cmidrule(lr){5-6}
Pair & $D$ & $k_{50}$ & $k_{80}$ & $k_{50}$ & $k_{80}$ & $\mathrm{CE}_{\mathrm{full}}$ \\
\midrule
Airline/Milk, 5 lags        & 5   & 24 & 32 & 24 & 32 & 0.89 \\
ETTh1/ETTh2, levels         & 6   & 24 & 64 & 24 & 48 & 0.74 \\
ETTh1/ETTh2, differenced    & 6   & 24 & 48 & 32 & 48 & 0.75 \\
Energy heating/cooling      & 8   & 32 & 64 & 32 & 48 & 0.58 \\
Energy, permuted target     & 8   & 16 & 24 & 32 & 48 & 0.94 \\
Wine red/white              & 11  & 32 & 48 & 32 & 48 & 0.86 \\
Airline/Milk, 20 lags       & 20  & 32 & 48 & 32 & 64 & 0.75 \\
Gas drift, batch 1/10       & 127 & 48 & 64 & 48 & 96 & 0.91 \\
ISOLET, speaker-disjoint    & 499 & 64 & 96 & 48 & 96 & 0.72 \\
\bottomrule
\end{tabular}
\end{table}

\begin{table}[h]
\centering
\small
\caption{Block 12: fraction of the full-replacement effect for the top-,
random- and bottom-$k$ subspaces. Mean over 24 draws, $\pm$ standard error
for the top subspace.}
\label{tab:real_b12}
\begin{tabular}{lrcccccc}
\toprule
Pair & $D$ & top-32 & rand-32 & bot-32 & top-64 & rand-64 & bot-64 \\
\midrule
Airline/Milk, 5 lags     & 5   & $0.864 \pm 0.010$ & 0.192 & 0.003 & $0.948 \pm 0.010$ & 0.347 & 0.003 \\
ETTh1/ETTh2, levels      & 6   & $0.694 \pm 0.019$ & 0.077 & 0.002 & $0.875 \pm 0.022$ & 0.128 & 0.005 \\
ETTh1/ETTh2, diff.       & 6   & $0.756 \pm 0.025$ & 0.106 & 0.004 & $0.886 \pm 0.025$ & 0.159 & 0.005 \\
Energy heating/cooling   & 8   & $0.641 \pm 0.043$ & 0.047 & 0.002 & $0.849 \pm 0.048$ & 0.057 & 0.007 \\
Energy, permuted         & 8   & $0.862 \pm 0.006$ & 0.038 & 0.001 & $0.928 \pm 0.007$ & 0.171 & 0.003 \\
Wine red/white           & 11  & $0.693 \pm 0.024$ & 0.080 & 0.002 & $0.896 \pm 0.019$ & 0.201 & 0.003 \\
Airline/Milk, 20 lags    & 20  & $0.513 \pm 0.028$ & 0.194 & 0.004 & $0.873 \pm 0.028$ & 0.305 & 0.008 \\
Gas drift, batch 1/10    & 127 & $0.183 \pm 0.029$ & 0.078 & 0.001 & $0.868 \pm 0.016$ & 0.099 & 0.004 \\
ISOLET                   & 499 & $0.209 \pm 0.043$ & 0.078 & 0.004 & $0.590 \pm 0.037$ & 0.113 & 0.015 \\
\bottomrule
\end{tabular}
\end{table}

\begin{table}[h]
\centering
\small
\caption{Block 24 (final block), as in Table~\ref{tab:real_b12}.}
\label{tab:real_b24}
\begin{tabular}{lrcccccc}
\toprule
Pair & $D$ & top-32 & rand-32 & bot-32 & top-64 & rand-64 & bot-64 \\
\midrule
Airline/Milk, 5 lags     & 5   & $0.860 \pm 0.007$ & 0.109 & 0.001 & $0.959 \pm 0.010$ & 0.344 & 0.003 \\
ETTh1/ETTh2, levels      & 6   & $0.715 \pm 0.019$ & 0.148 & 0.001 & $0.943 \pm 0.023$ & 0.268 & 0.002 \\
ETTh1/ETTh2, diff.       & 6   & $0.614 \pm 0.019$ & 0.208 & 0.002 & $0.897 \pm 0.024$ & 0.385 & 0.007 \\
Energy heating/cooling   & 8   & $0.631 \pm 0.036$ & 0.174 & 0.001 & $0.910 \pm 0.052$ & 0.321 & 0.002 \\
Energy, permuted         & 8   & $0.738 \pm 0.006$ & 0.029 & 0.000 & $0.949 \pm 0.007$ & 0.126 & 0.000 \\
Wine red/white           & 11  & $0.670 \pm 0.019$ & 0.155 & 0.002 & $0.925 \pm 0.018$ & 0.296 & 0.001 \\
Airline/Milk, 20 lags    & 20  & $0.617 \pm 0.012$ & 0.121 & 0.000 & $0.888 \pm 0.025$ & 0.340 & 0.002 \\
Gas drift, batch 1/10    & 127 & $0.315 \pm 0.011$ & 0.064 & 0.001 & $0.770 \pm 0.017$ & 0.232 & 0.002 \\
ISOLET                   & 499 & $0.251 \pm 0.015$ & 0.168 & 0.006 & $0.770 \pm 0.029$ & 0.362 & 0.020 \\
\bottomrule
\end{tabular}
\end{table}

\section{Kernel Decoder}
\label{app:decoder_all}
This appendix supports Section~\ref{sec:decoder}; its subsections follow the order of that section.

\subsection{Identifiability of the Spectral Density}
\label{app:identifiability}

We characterize what is fundamentally retrievable about the spectral density
$S(\omega)$ of a stationary GP from observed function data, in two regimes:
single-realization and multi-realization.

\subsubsection{Single-Realization Limit}
\begin{theorem}[Single-function non-identifiability of spectral weights]
\label{thm:single-nonident}
Let $f \sim \mathcal{GP}(0, k)$ with continuous stationary kernel and spectral
density  $S(\omega)=\sum_{q=1}^{Q}\frac{w_q}{2}\Big[\mathcal{N}\!\left(\omega \mid \mu_q,\sigma_q^2\right)+\mathcal{N}\!\left(\omega \mid -\mu_q,\sigma_q^2\right)\Big],\; w_q>0$. Let $\{f(x_i)\}_{i=1}^{N}$ be a single realization on a fixed grid,
normalized to unit empirical variance. Then $\{w_q\}_{q=1}^{Q}$ is not
identifiable from this single realization, except up to a common multiplicative
constant, even in the limit $N \to \infty$.
\end{theorem}

\begin{proof}
Let $\mathbf{f} = (f(x_1),\dots,f(x_N))^\top$ denote the vector of observations.
Since $f \sim \mathcal{GP}(0,k)$ is stationary and Gaussian,
\[
\mathbf{f} \sim \mathcal{N}(0, K),
\qquad
K_{ij} = k(x_i - x_j).
\]

By Bochner's theorem, the kernel admits the spectral representation
\[
k(\tau) = \int_{\mathbb{R}} e^{2\pi i \omega \tau} S(\omega)\, d\omega,
\qquad
   S(\omega)=\sum_{q=1}^{Q}\frac{w_q}{2}\Big[\mathcal{N}\!\left(\omega \mid \mu_q,\sigma_q^2\right)+\mathcal{N}\!\left(\omega \mid -\mu_q,\sigma_q^2\right)\Big],\; w_q>0.
\]

Consider any constant $c > 0$ and define a rescaled spectral density
\[
\tilde S(\omega) = c \, S(\omega),
\]
with corresponding kernel
\[
\tilde k(\tau) = c \, k(\tau).
\]
Let $\tilde f \sim \mathcal{GP}(0,\tilde k)$. Then $\tilde f$ and $f$ are related
in distribution by
\[
\tilde f(x) \stackrel{d}{=} \sqrt{c}\, f(x).
\]

Let $\tilde{\mathbf{f}} = (\tilde f(x_1),\dots,\tilde f(x_N))^\top$.
Under empirical variance normalization which normalization mirrors the preprocessing used in PFNs, making the model invariant to global rescaling of function values.

\[
\hat{\mathbf{f}}
= \frac{\mathbf{f} - \bar f \mathbf{1}}{\|\mathbf{f} - \bar f \mathbf{1}\|_2},
\qquad
\hat{\tilde{\mathbf{f}}}
= \frac{\tilde{\mathbf{f}} - \overline{\tilde f} \mathbf{1}}
{\|\tilde{\mathbf{f}} - \overline{\tilde f} \mathbf{1}\|_2}.
\]
Since $\tilde{\mathbf{f}} = \sqrt{c}\, \mathbf{f}$, it follows immediately that
\[
\hat{\tilde{\mathbf{f}}} = \hat{\mathbf{f}}.
\]

Therefore, the normalized single-sample distribution induced by the kernel $k(\tau)$ is identical to that induced by $\tilde k(\tau) = c k(\tau)$. Because scaling the spectral density corresponds exactly to scaling all weights
$\{w_q\}$ by the same constant $c$, no estimator operating on a single normalized realization can distinguish between $\{w_q\}$ and $\{c w_q\}$.Consequently, the spectral weights are not identifiable from a single realization,
except up to a common multiplicative constant, even as $N \to \infty$.
\end{proof}

\paragraph{Remark.}
Frequency identifiability follows from classical spectral estimation theory:
the discrete Fourier transform of a stationary process concentrates energy near the true frequencies, while global rescaling of the covariance affects only the
magnitude, not the location, of spectral peaks.

\begin{proposition}[Unbiased kernel-scale estimator]
\label{prop:scale-estimator}
Let $\mathbf{f}\sim\mathcal{N}(0, \alpha K)$ for known PSD matrix $K$ and unknown $\alpha>0$. Then $\hat\alpha = \|\mathbf{f}\|_2^{2}/\mathrm{tr}(K)$ satisfies $\mathbb{E}[\hat\alpha]=\alpha$.
\end{proposition}

\begin{proof}
Let $\mathbf{f} = (f(x_1),\dots,f(x_N))^\top \in \mathbb{R}^N$ denote the vector of
function values evaluated at the fixed input locations $\{x_i\}_{i=1}^N$.
Consider,
\[
\mathbf{f} \sim \mathcal{N}(0, \alpha K_{\text{pred}}),
\]
where $K_{\text{pred}} \in \mathbb{R}^{N\times N}$ is a fixed positive semi-definite matrix predicited using a single realization decoder and $\alpha>0$ is an unknown scalar.

We define the squared $\ell_2$ norm of $\mathbf{f}$ as
$\|\mathbf{f}\|_2^2 = \mathbf{f}^\top \mathbf{f}$.
All expectations below are taken with respect to the randomness of
$\mathbf{f}$ induced by the Gaussian process, i.e.
$\mathbb{E}[\cdot] = \mathbb{E}_{\mathbf{f} \sim \mathcal{N}(0,\alpha K_{\text{pred}})}[\cdot]$.

Since $\mathbf{f}$ is zero-mean Gaussian,
\[
\mathbb{E}[\mathbf{f}\mathbf{f}^\top] = \alpha K_{\text{pred}}.
\]
Taking the trace on both sides yields
\[
\mathbb{E}[\|\mathbf{f}\|_2^2]
=
\mathbb{E}[\operatorname{tr}(\mathbf{f}\mathbf{f}^\top)]
=
\operatorname{tr}(\mathbb{E}[\mathbf{f}\mathbf{f}^\top])
=
\alpha \operatorname{tr}(K_{\text{pred}}).
\]

Therefore, for the estimator
\[
\hat{\alpha} \;=\; \frac{\|\mathbf{f}\|_2^2}{\operatorname{tr}(K_{\text{pred}})},
\]
we obtain
\[
\mathbb{E}[\hat{\alpha}]
=
\frac{\mathbb{E}[\|\mathbf{f}\|_2^2]}{\operatorname{tr}(K_{\text{pred}})}
=
\alpha,
\]
which proves that $\hat{\alpha}$ is an unbiased estimator of the kernel scale.
\end{proof}

\subsubsection{Multi-Realization Guarantee}
\begin{theorem}[Identifiability from multiple realizations]
\label{thm:multi-ident}
   Let $\{f_m\}_{m=1}^M$ be i.i.d.\ realizations from a zero-mean stationary GP with
   spectral density $S(\omega)$ given in Section 2.2, $w_q>0$, $\mu_q\ge 0$ and
   distinct pairs $(\mu_q,\sigma_q)$, observed on a common fixed grid. As $M\to\infty$, the
   empirical covariance converges to $k(\tau)$ in probability, and $\{w_q\}_{q=1}^Q$
   becomes identifiable from second-order statistics, up to permutation of components.\end{theorem} 

\begin{proof}
We proceed by showing that multiple independent realizations allow consistent estimation of the covariance function, which uniquely determines the spectral
weights.

For a zero-mean stationary Gaussian process, the covariance function
\[
k(\tau) = \mathbb{E}[f(x) \cdot f(x+\tau)]
\]
fully characterizes the process. By Bochner’s theorem~\cite{gpml}, the covariance function $k(\tau)$ is in one-to-one correspondence with the spectral density $S(\omega)$. Therefore, identifying $k(\tau)$ is equivalent to identifying $S(\omega) = \sum_{q=1}^Q w_q \, \mathcal{N}(\omega \mid \mu_q, \sigma_q^2)$ and its parameters $\{w_q,\mu_q,\sigma_q\}$. We thus show that the empirical covariance converges to $k(\tau)$ as the number of realizations $M$ increases.

Fix any pair of input locations $(x_i,x_j)$ and define the empirical covariance estimator across realizations:
\[
\hat{k}_M(x_i,x_j)
\;\triangleq\;
\frac{1}{M} \sum_{m=1}^M \big[f_m(x_i) \cdot f_m(x_j)\big].
\]
Since the realizations $\{f_m\}$ are independent and identically distributed, each term $f_m(x_i) \cdot f_m(x_j)$ is an independent sample of a random variable
with expectation
\[
\mathbb{E}\!\left[f_m(x_i) \cdot f_m(x_j)\right]
=
k(x_i-x_j),
\]
where the expectation is taken with respect to the Gaussian process prior.

Moreover, because $f_m(x_i) \cdot f_m(x_j)$ has finite second moment under the Gaussian process prior~\cite{gpml}, the Law of Large Numbers implies
\[
\hat{k}_M(x_i,x_j)
\xrightarrow{P}
k(x_i-x_j)
\quad \text{as } M \to \infty.
\]
Since the input grid $\{x_i\}_{i=1}^N$ is fixed and finite, this convergence holds jointly for all pairs $(i,j)$. Consequently, the entire empirical covariance matrix converges in probability to the true covariance matrix:
\[
\big[\hat{k}_M(x_i,x_j)\big]_{i,j=1}^N
\xrightarrow{P}
\big[k(x_i-x_j)\big]_{i,j=1}^N.
\]

Finally, the limiting covariance function $k(\tau)$ uniquely determines the spectral density $S(\omega)$ via Bochner’s theorem. In particular, for the spectral mixture form
\[
S(\omega) = \sum_{q=1}^Q w_q \, \mathcal{N}(\omega \mid \mu_q, \sigma_q^2),
\]
the parameters $\{w_q\}$ are uniquely determined by $k(\tau)$. Therefore, as the number of independent realizations $M$ increases, the spectral weights $\{w_q\}$ become identifiable from the empirical second-order statistics of the observed functions.
\end{proof}

\subsection{Decoder Architecture and Training}
\label{app:decoder}

\subsubsection{Pipeline Overview}
The decoder takes a context set $\mathcal{D}_{\text{ctx}}$, processes it
through the frozen PFN to obtain $H, V$, and outputs explicit spectral
parameters from which a stationary kernel is reconstructed via Bochner's
theorem. The PFN is never updated.

\subsubsection{Multi-Query Attention Pooling}
For an input sequence $H \in \mathbb{R}^{B \times N \times d}$ and learned queries $Q \in \mathbb{R}^{1 \times n_q \times d}$,
\[
\mathrm{MQA}(H) \;=\; \mathrm{LinearFlatten}\bigl(\mathrm{MultiheadAttn}(Q, H, H)\bigr) \in \mathbb{R}^{B\times d}.
\]
Independent MQA modules are applied to $H$ and $V$, giving $z_H = \mathrm{MQA}_H(H)$, $z_V = \mathrm{MQA}_V(V)$, fused as $z = \mathrm{MLP}([z_H \,\Vert\, z_V])$.

\subsubsection{Spectral Parameter Heads}
We discretize the frequency range $[\mu_{\min},\mu_{\max}]$ into $B$ bins of
width $\Delta = (\mu_{\max}-\mu_{\min})/B$. Three heads predict, per bin:
(i) activation probability $p_b$, (ii) offset and bandwidth $(\delta_b,\sigma_b)$
giving $\mu_b = \mu_{\min} + (b+\delta_b)\Delta$, and (iii) weight $w_b$
(multi-realization only; the single-realization decoder uses uniform $w_b=1$
following Theorem~\ref{thm:single-nonident}).

\subsubsection{Kernel Reconstruction}
\begin{equation}
K(\tau) \;=\; \sum_{b:\,p_b > \gamma}\! w_b \exp\!\bigl(-2\pi^2 \sigma_b^2 \tau^2\bigr)\cos(2\pi \mu_b \tau),
\qquad K_{\text{pred}} \;=\; \hat\alpha\, K,
\label{eq:kernel-reconstruction}
\end{equation}
with classification threshold $\gamma$ and analytical scale
$\hat\alpha = \|\mathbf{f}\|_2^2/\mathrm{tr}(K)$.

\subsubsection{Loss and Curriculum}
The decoder is trained with a composite loss
\begin{equation}
\mathcal{L} \;=\; \mathcal{L}_{\text{BCE}}(p, y_{\text{bin}}) \;+\; \lambda \!\sum_{b\in\text{active}}\! \|\theta_b - \theta^{*}_b\|^2,
\label{eq:decoder-loss}
\end{equation}
with $w_{\text{pos}}=30$ in BCE to handle sparse positive bins, and a curriculum
that ramps $n_p$ (active components) from 1 to 4. Hyperparameters in
Table~\ref{tab:decoder-hyperparams}.

\begin{table}[h]
\centering\small
\caption{Decoder training hyperparameters.}
\label{tab:decoder-hyperparams}
\begin{tabular}{lcc}
\toprule
Parameter & Multi-Realization & Single-Realization \\
\midrule
$n_{\text{samples}}$            & 100{,}000 & 300{,}000 \\
$n_{\text{points}}$             & 200       & 200 \\
$n_{\text{bins}}$               & 50        & 50 \\
$d_{\text{model}}$              & 128       & 128 \\
$d_{\text{ff}}$                 & 256       & 256 \\
$n_{\text{queries}}$            & 4         & 4 \\
dropout                         & 0.1       & 0.1 \\
BCE positive weight             & 30.0      & 30.0 \\
$\lambda_{\text{reg}}$          & 5.0       & 5.0 \\
learning rate                   & $10^{-3}$ & $10^{-3}$ \\
weight decay                    & $10^{-4}$ & $10^{-4}$ \\
GP samples per task ($M$)       & 16        & 1 \\
Frequency range (Hz)            & $[0.5, 3.0]$ & $[0.5, 3.0]$ \\
$\sigma$ range                  & $[0.01, 0.05]$ & $[0.01, 0.05]$ \\
Epochs (Phases 1 / 2 / 3)       & 1000 / 1000 / 2000 & 200 / 200 / 400 \\
\bottomrule
\end{tabular}
\end{table}

\subsubsection{Training Data Generation}
\paragraph{Multi-Realization.}
Spectral parameters: $\mu_q\sim\mathcal{U}[\mu_{\min},\mu_{\max}]$,
$\sigma_q\sim\mathcal{U}[\sigma_{\min},\sigma_{\max}]$,
$w_q\sim\mathrm{Gamma}(2,1)$. Kernel
$K=\sum_q w_q \exp(-2\pi^2\sigma_q^2\tau^2)\cos(2\pi\mu_q\tau)$. GP samples
$y^{(m)}\!\sim\!\mathcal{N}(0,K)$, $m=1,\dots,M$.

\paragraph{Single-Realization.}
Random Fourier Feature (RFF) signals
\begin{equation}
y(x) \;=\; \sqrt{\tfrac{2}{n_{\text{rff}} \cdot n_p}} \sum_{q=1}^{n_p}\sum_{j=1}^{n_{\text{rff}}}\!\cos(2\pi\omega_{qj} x + \phi_{qj}),
\label{eq:rff-signal}
\end{equation}
with $\omega_{qj}\!\sim\!\mathcal{N}(\mu_q,\sigma_q^2)$,
$\phi_{qj}\!\sim\!\mathcal{U}[0,2\pi]$, $n_{\text{rff}}=100$.

\subsection{Decoder Controls}
\label{app:decoder_controls}
All variants in Table~\ref{tab:decoder_controls} use the four kernel families of Table~\ref{tab:spectral-baselines} and 5 seeds. For the random-PFN and single-stream rows, the decoder trained on DVA-PFN latents is evaluated with only its input changed: latents of a randomly initialized DVA-PFN of the same architecture, or the DVA-PFN latents with the $V$ (respectively $H$) stream removed. The raw-input decoder replaces the frozen PFN with a trainable encoder of the raw $(x,y)$ context and is trained end-to-end with the identical Filter Bank head, prior and schedule, so it optimizes more parameters than the latent decoder. It reaches Ker-CKA within 0.012--0.016 of the latent decoder on the periodic families and a lower final training loss (BCE 0.7145 vs.\ 0.7678). Two of its three training restarts collapsed to a constant solution; we report the best restart.

\begin{table}[h]
\centering\small
\caption{Decoder controls: Ker-MSE ($\times10^{-3}$, $\downarrow$), mean over 5 seeds. LP: locally periodic.}
\label{tab:decoder_controls}
    \begin{tabular}{lcccc}
    \toprule
    Variant & RBF & Periodic & LP & SM \\
    \midrule
    Full decoder  & 6.4   & 3.0   & 1.3   & 0.9 \\
    Raw input     & 6.1   & 3.1   & 1.2   & 0.9 \\
    Random PFN    & 518.6 & 346.0 & 229.6 & 146.7 \\
    $H$ only      & 6.3   & 3.0   & 1.3   & 0.9 \\
    $V$ only      & 572.2 & 349.6 & 256.6 & 187.6 \\
    \bottomrule
    \end{tabular}
\end{table}

\subsection{Baselines and Predictive Results}
\label{app:baselines}

\textbf{Deep Kernel Learning (DKL).}
Feature extractor: 3-layer MLP ($1\!\to\!64\!\to\!64\!\to\!2$), Tanh hidden activations, linear output layer. GP: \texttt{ScaleKernel(RBFKernel(ard\_num\_dims=2))} on the 2D latent. Joint MLP and GP optimization via Adam (lr$=0.01$, 500 iterations, exact MLL). 

\textbf{Random Fourier Features (RFF).}
Feature map: $\phi(x) = \sqrt{2/n}\cos(xW^\top/\ell + b)$, $n=128$. $W\!\sim\!\mathcal{N}(0,I)$ and $b\!\sim\!\mathcal{U}[0,2\pi]$ are fixed (non-optimized) buffers. Learnable: softplus-constrained $\ell$, output scale, noise variance. GP: \texttt{ScaleKernel(LinearKernel())} on $\phi(x)$. Same optimizer, budget, and standardization as DKL.

\textbf{GP Oracle.}
Exact GP with ground-truth kernel family, \texttt{GaussianLikelihood}, and \texttt{SpectralMixtureKernel} initialized via \texttt{initialize\_from\_data} for SM-$Q$ tasks. Five restarts per task (Adam, lr$=0.1$, 500 iterations); lowest MLL selected.

\begin{figure}[h]
\centering
\includegraphics[width=\linewidth]{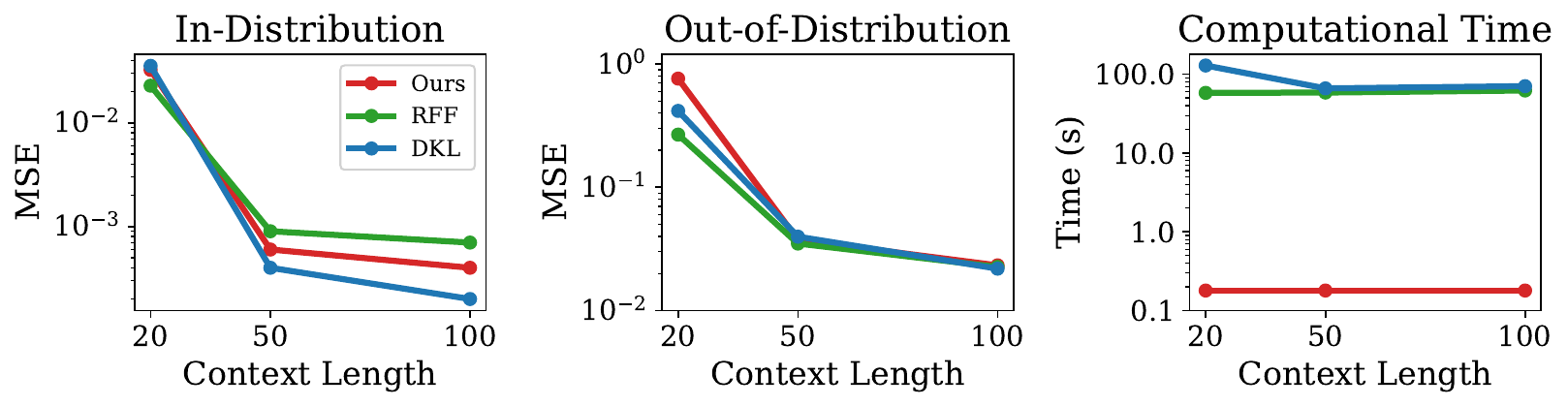}
\caption{GP-MSE versus context length on (left) in-distribution sinusoids and (centre) out-of-distribution triangle waves with random frequencies $\sim\mathcal{U}(1.0, 3.0)$. Decoded kernels match DKL/RFF on both, without per-task optimization. Computational time (right) is reported over 100 tasks.}
\label{fig:ood_context_scaling}
\end{figure}

\begin{figure}[h]
\centering
\includegraphics[width=0.80\linewidth]{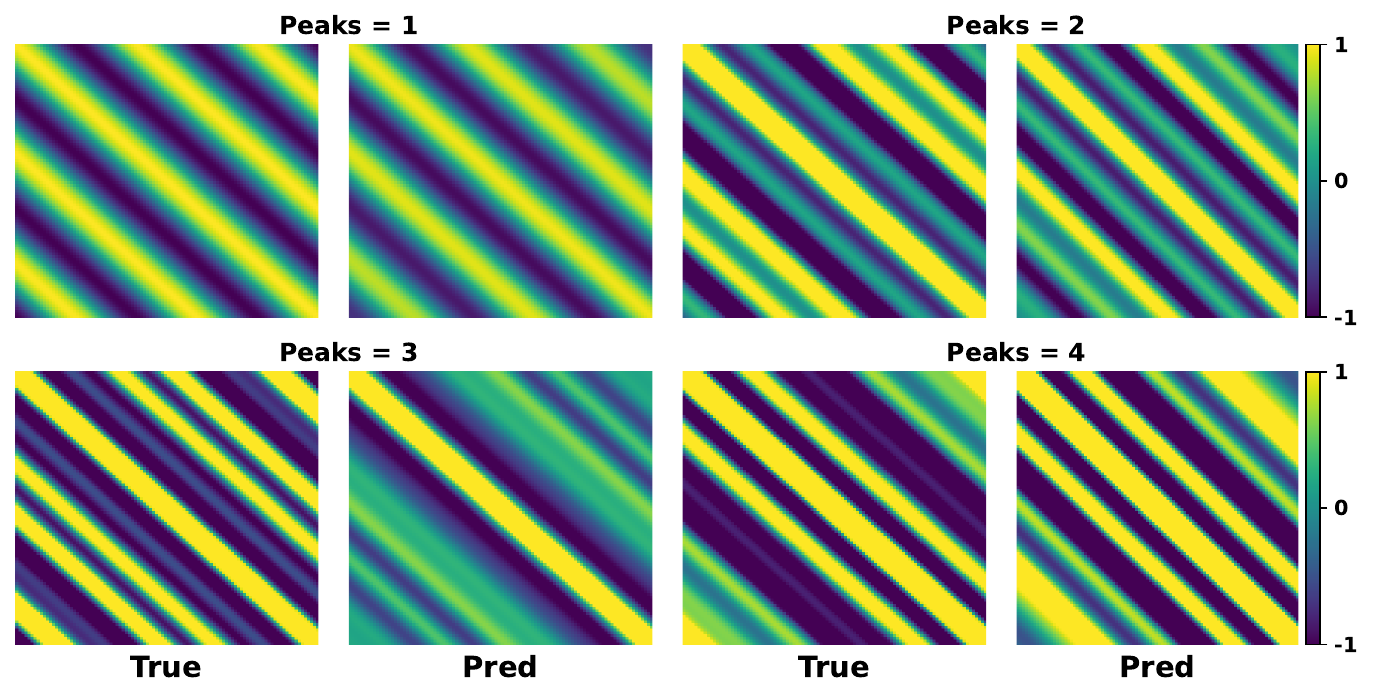}
\caption{Kernel reconstruction from \textbf{single function observations}. Global amplitude is ambiguous (Theorem~\ref{thm:single-nonident}); dominant periodic structure and lengthscales are recovered.}
\label{fig:kernel_reconstruction_single}
\end{figure}

\begin{figure}[h]
\centering
\includegraphics[width=0.45\linewidth]{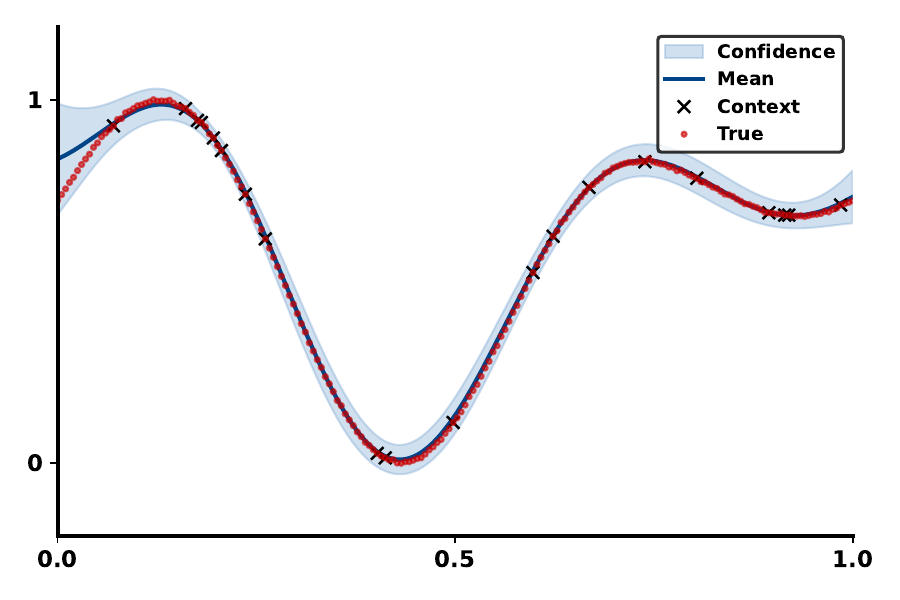}
\hfill
\includegraphics[width=0.45\linewidth]{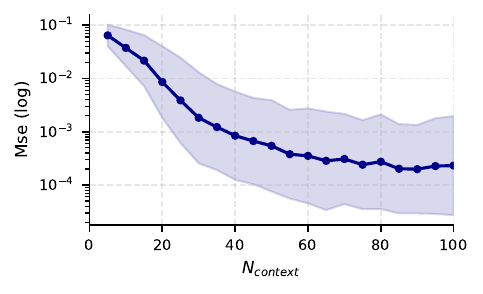}
\caption{\textbf{Single-realization} decoder performance. (a) Qualitative GP regression from a decoded kernel with $N=20$ context points. (b) GP-MSE versus context size $N_{\text{context}}$. The gap to the oracle remains roughly constant in $N$, consistent with the single-realization weight ambiguity (Theorem~\ref{thm:single-nonident}).}
\label{fig:single-realization-scaling}
\end{figure}

Table~\ref{tab:nlpd} gives the predictive log-density of the decoded kernel against the baselines of Table~\ref{tab:cookbook}. Figure~\ref{fig:ood_context_scaling} varies the context length in and out of distribution, and Figures~\ref{fig:kernel_reconstruction_single} and~\ref{fig:single-realization-scaling} show decoded kernels and single-realization GP regression.

\begin{table}[h]
\centering\small
\caption{NLPD ($\downarrow$) on 150 held-out targets with 50 context points; mean $\pm$ std over 100 realizations. The decoder uses DVA-PFN.}
\label{tab:nlpd}
    \begin{tabular}{lcccc}
    \toprule
    Kernel & Decoder & DKL & RFF & Oracle GP \\
    \midrule
    RBF             & $-2.64 \pm 0.58$ & $-1.63 \pm 0.19$ & $-1.62 \pm 0.17$ & $-2.86 \pm 0.45$ \\
    Periodic        & $-1.66 \pm 0.84$ & $-1.45 \pm 0.37$ & $-0.79 \pm 0.94$ & $-1.94 \pm 1.50$ \\
    $K_{\text{RP}}$ & $-1.53 \pm 0.96$ & $-1.52 \pm 0.21$ & $-0.71 \pm 0.96$ & $-2.86 \pm 0.34$ \\
    SM ($Q=2$)      & $-2.67 \pm 0.67$ & $-1.64 \pm 0.16$ & $-1.65 \pm 0.20$ & $-3.08 \pm 0.54$ \\
    SM ($Q=4$)      & $-2.27 \pm 1.38$ & $-1.63 \pm 0.06$ & $-1.12 \pm 1.11$ & $-3.00 \pm 0.46$ \\
    \bottomrule
    \end{tabular}
\end{table}

\subsection{Comparison with Classical Spectral Estimation}
\label{app:classical}
\begin{figure}[h]
\centering
\includegraphics[width=\linewidth]{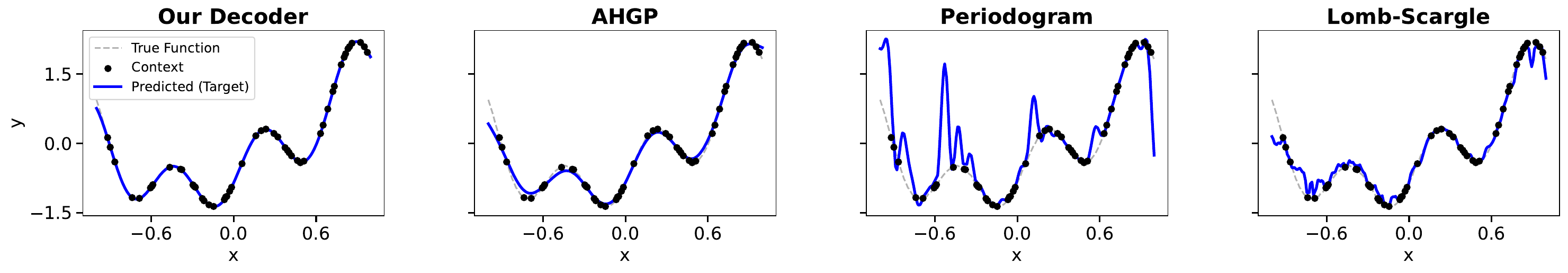}
\caption{GP predictive posterior on a complex spectral mixture from a singlerealization. Periodogram and Lomb--Scargle interpolate the 50 context points exactly but oscillate wildly between observations; the decoded kernel recovers the underlying structure and tracks the held-out target.}
\label{fig:posterior_comparison}
\end{figure}

We fit spectral mixtures to the standard and Lomb--Scargle periodograms of each context, build kernels via Bochner's theorem, and use them for GP regression on held-out targets (Table~\ref{tab:spectral-baselines}, Figure~\ref{fig:posterior_comparison}).

\paragraph{Estimators.}
\begin{align}
\text{Periodogram: } && P(\omega) &= \tfrac{1}{N}\Bigl|\textstyle\sum_{n=1}^{N} y_n e^{-i\omega x_n}\Bigr|^2, \\
\text{Lomb--Scargle: } && P_{\text{LS}}(\omega) &= \tfrac{1}{2\sigma^2}\!\left[\tfrac{(\sum_n y_n\cos\omega(x_n-\tau))^2}{\sum_n \cos^2\omega(x_n-\tau)} + \tfrac{(\sum_n y_n\sin\omega(x_n-\tau))^2}{\sum_n \sin^2\omega(x_n-\tau)}\right],
\end{align}
where $\tau$ is the standard frequency-dependent time delay.

\begin{table}[h]
\centering\small
\caption{Spectral recovery benchmark (50 context, 150 target points).
Classical methods overfit the context set (low Ker-MSE) but fail at GP-MSE on
unseen targets. The PFN-based decoder uses pretraining-induced regularization
to generalize. We report Mean MSE over 100 samples. \textbf{Bold} marks best per column within each kernel family.}
\label{tab:spectral-baselines}
\begin{tabular}{llccc}
\toprule
Kernel family & Method & Ker-MSE ($\downarrow$) & Ker-CKA ($\uparrow$) & GP-MSE ($\downarrow$) \\
\midrule
\multirow{4}{*}{RBF}
& Periodogram        & 0.0060          & \textbf{0.7316} & 0.0859 \\
& Lomb--Scargle      & \textbf{0.0031} & 0.7177          & 0.0788 \\
& PFN Decoder (Ours) & 0.0062          & 0.3648          & 0.0011 \\
\midrule
\multirow{4}{*}{Periodic}
& Periodogram        & 0.0028          & \textbf{0.8768} & 0.0964 \\
& Lomb--Scargle      & \textbf{0.0023} & 0.7191          & 0.0468 \\
& PFN Decoder (Ours) & 0.0031          & 0.5791          & \textbf{0.0015} \\
\midrule
\multirow{4}{*}{Locally Periodic}
& Periodogram        & 0.0012          & \textbf{0.7730} & 0.0711 \\
& Lomb--Scargle      & \textbf{0.0008} & 0.7118          & 0.0467 \\
& PFN Decoder (Ours) & 0.0012          & 0.6378          & \textbf{0.0027} \\
\midrule
\multirow{4}{*}{Spectral Mixture}
& Periodogram        & 0.0008          & \textbf{0.7870} & 0.1174 \\
& Lomb--Scargle      & \textbf{0.0005} & 0.7676          & 0.0989 \\
& PFN Decoder (Ours) & 0.0008          & 0.7007          & \textbf{0.0017} \\
\bottomrule
\end{tabular}
\end{table}

\paragraph{Why classical methods fail at generalization.}
Classical estimators achieve very low Ker-MSE on the context set --- they memorize the discrete-sample artifacts, noise, and phase alignments of the specific realization. Theorem~\ref{thm:single-nonident} predicts this: the
weights are ambiguous from a single realization, and least-squares fits seize on whatever assignment best matches the observed samples. Once tested on
unseen target locations, this overfit collapses, producing GP-MSE that is
$17$--$78\times$ worse than the PFN decoder across all four families. The
decoder sidesteps this by inheriting the structural prior the PFN learned
during pretraining: peak locations and bandwidths come from a frozen
representation that has already integrated over many realizations, so they
generalize to new target points rather than tracking the context.

\subsection{Bayesian Optimization}
\label{app:bo}
The decoded kernel parameterizes a standard GP that can be evaluated and
updated without re-invoking the PFN. We illustrate this with a Bayesian
optimization pipeline: PFN $\to$ decoder $\to$ SM kernel $\to$ GP posterior
$\to$ UCB acquisition. The PFN runs once on the initial context to extract
$k(\tau)$; subsequent BO iterations use only the decoded kernel.

\paragraph{Setup.} 100 objective functions per kernel type (SM-$Q1$,
SM-$Q2$, SM-$Q4$); $N_0 = 15$ initial observations; 50 BO iterations; UCB
acquisition. Final simple regret is reported in Table~\ref{tab:bo_regret}.

\begin{table}[h]
\centering\small
\caption{Bayesian optimization with the decoded kernel. Final simple regret (mean $\pm$ std, $\downarrow$ better, 100 functions each).}
\label{tab:bo_regret}
\begin{tabular}{lcc}
\toprule
Kernel type & Decoded Kernel + GP & Oracle GP \\
\midrule
SM ($Q\!=\!1$) & $0.001\pm0.002$ & $0.005\pm0.014$ \\
SM ($Q\!=\!2$) & $0.002\pm0.009$ & $0.004\pm0.015$ \\
SM ($Q\!=\!4$) & $0.006\pm0.029$ & $0.004\pm0.028$ \\
\bottomrule
\end{tabular}
\end{table}

\begin{table}[h]
\centering\small
\caption{Resource comparison for iterative inference on CPU.}
\label{tab:bo_resources}
\begin{tabular}{lcc}
\toprule
& PFN (per step) & Decoded Kernel + GP \\
\midrule
Parameters             & 859{,}108 & 9 \\
Memory                 & 3.4 MB    & 72 B \\
Time / iteration (CPU) & 24.8 ms   & 4.8 ms \\
GPU required           & yes       & no \\
\bottomrule
\end{tabular}
\end{table}

\paragraph{What this shows.} The decoded kernel matches the oracle on simple mixtures and stays within one order of magnitude on $Q=4$, while the entire BO loop runs on CPU at $\sim$5\,ms per step. A PFN can serve as a BO surrogate \citep{muller2022bayesian}, but each step then requires a GPU forward pass over the full context; the decoded kernel gives a closed-form posterior that is updated incrementally on CPU, validating the practical claim that the recovered Bayesian object is not just inspectable but reusable.

\subsection{Multi-Realization Decoder and Higher Dimensions}
\label{app:decoder-results}\label{app:high-dim-protocol}\label{app:scaling}
\begin{figure}[h]
\centering
\includegraphics[width=0.6\linewidth]{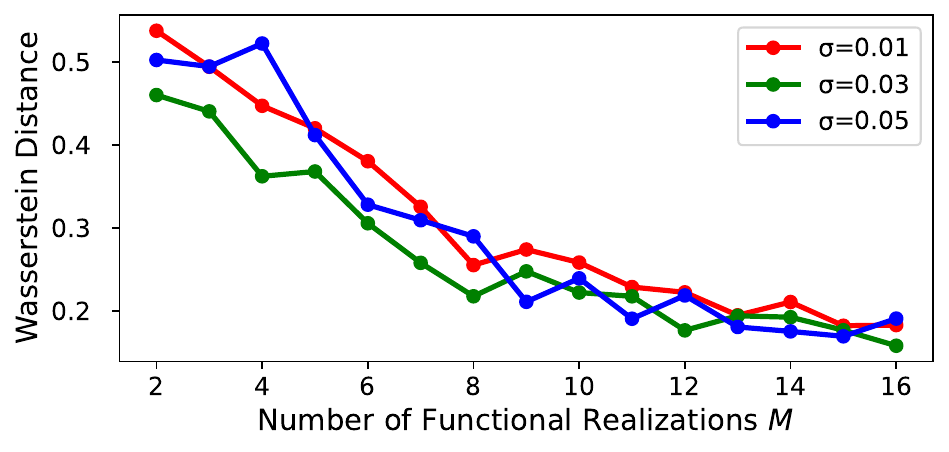}
\caption{Wasserstein distance between true and decoded spectral densities as a function of the number of independent realizations $M$, for three bandwidths $\sigma\in\{0.01, 0.03, 0.05\}$. The overall decrease is consistent with Theorem~\ref{thm:multi-ident}.}
\label{fig:wasserstein_M}
\end{figure}

\begin{figure}[h]
\centering
\includegraphics[width=0.7\linewidth]{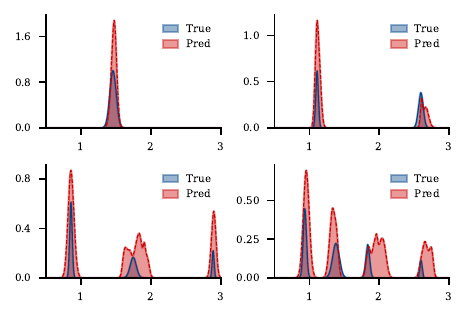}

\caption{Decoded vs.\ ground-truth spectral densities for 1--4 component mixtures.}
\label{fig:spectral_den_recover}
\end{figure}

Figures~\ref{fig:wasserstein_M} and~\ref{fig:spectral_den_recover} show spectral recovery as the number of realizations grows, and Tables~\ref{tab:ood-kernels} and~\ref{tab:high-dim} give GP-MSE for out-of-distribution kernel families and for 5D and 10D additive kernels, generated as follows.

Functions are generated from additive GP priors
$K(x,x') = \frac{1}{|\mathcal{D}|}\sum_{d\in\mathcal{D}} K_d(x_d,x_d')$, where $\mathcal{D}$ is the set of active dimensions and each $K_d$ is a 1D kernel (RBF, periodic, or spectral mixture). For each function only 2--4 dimensions are active in 5D and up to 6 in 10D, ensuring well-conditioned
covariance matrices. Inputs $x_d \in [0,1]$ are sampled uniformly per dimension and sorted before kernel evaluation, removing combinatorial spatial variation while preserving the structure each $K_d$ induces. Multiplicative kernel composition is avoided in high dimensions because it collapses covariance structure.

The decoder architecture is unchanged across dimensions; in 10D we use a
four-phase curriculum on the number of active dimensions
($2\!\to\!3\!\to\!4\!\to\!\leq\!6$). PFN training in 5D and 10D uses the
same architecture and hyperparameters as in 1D, with additive priors only.

\begin{table}[h]
\centering
\caption{Decoder GP-MSE on a fixed support of 16 functions per kernel, evaluated on 20 unseen test functions from the same prior. RBF and Mat\'ern families lie outside the sparse-spectral-mixture model class; the decoder shows graceful degradation rather than failure.}
\label{tab:ood-kernels}
\begin{tabular}{lcc}
\toprule
Kernel family & Oracle GP MSE & Decoder MSE \\
\midrule
SM ($Q\!=\!1$)     & $1.13\!\times\!10^{-4}$  & $6.15\!\times\!10^{-4}$ \\
SM ($Q\!=\!2$)     & $1.45\!\times\!10^{-4}$ & $1.14\!\times\!10^{-4}$ \\
SM ($Q\!=\!4$)     & $1.17\!\times\!10^{-4}$ & $2.99\!\times\!10^{-4}$ \\
\midrule
\multicolumn{3}{c}{\emph{Out-of-distribution}} \\
\midrule
RBF                & $1.34\!\times\!10^{-4}$ & $4.76\!\times\!10^{-3}$ \\
Mat\'ern-1/2       & $9.50\!\times\!10^{-2}$ & $1.78\!\times\!10^{-1}$ \\
Mat\'ern-3/2       & $1.14\!\times\!10^{-1}$ & $2.21\!\times\!10^{-1}$ \\
Mat\'ern-5/2       & $1.61\!\times\!10^{-1}$ & $2.94\!\times\!10^{-1}$ \\
\bottomrule
\end{tabular}
\end{table}

\begin{table}[h]
\centering\small
\caption{Multi-realization decoder GP-MSE in 5D and 10D additive-kernel settings, $M=16$ functions per task. Predictions remain within a small constant factor of the oracle except on Periodic, which is hardest under additive structure.}
\label{tab:high-dim}
\begin{tabular}{lcc}
\toprule
Kernel (additive)        & Oracle GP MSE & Decoder MSE \\
\midrule
RBF (5D)                 & $1.30\!\times\!10^{-4}$ & $2.90\!\times\!10^{-4}$ \\
SM ($Q\!=\!1$, 5D)       & $1.10\!\times\!10^{-4}$ & $3.00\!\times\!10^{-4}$ \\
SM ($Q\!=\!2$, 5D)       & $1.20\!\times\!10^{-4}$ & $2.40\!\times\!10^{-4}$ \\
SM ($Q\!=\!4$, 5D)       & $1.30\!\times\!10^{-4}$ & $2.90\!\times\!10^{-4}$ \\
Periodic (5D)            & $2.60\!\times\!10^{-4}$ & $9.50\!\times\!10^{-3}$ \\
\midrule
RBF (10D)                & $1.30\!\times\!10^{-4}$ & $3.25\!\times\!10^{-4}$ \\
SM ($Q\!=\!1$, 10D)      & $1.15\!\times\!10^{-4}$ & $4.01\!\times\!10^{-4}$ \\
SM ($Q\!=\!2$, 10D)      & $3.19\!\times\!10^{-4}$ & $5.38\!\times\!10^{-4}$ \\
SM ($Q\!=\!4$, 10D)      & $2.06\!\times\!10^{-4}$ & $4.56\!\times\!10^{-4}$ \\
Periodic (10D)           & $2.71\!\times\!10^{-4}$ & $2.32\!\times\!10^{-3}$ \\
\bottomrule
\end{tabular}
\end{table}

\section{Architecture Ablation Validating Mechanistic Predictions}
\label{sec:abl}

\paragraph{Setup.}
We sweep the architecture along three axes using DVA-PFN: width $d \in \{16, 32, 48, 64, 96, 128\}$, depth $L \in \{2, 3, 6\}$, and attention variant (joint multi-head Standard vs.\ multi-query MQA), training each of the 36 combinations with 3 random seeds for a total of 108 runs. All runs share the training prior and pre-processing of Appendix~\ref{app:pfn_pretraining} (hierarchical spectral mixture, sigmoid-normalized targets), the same optimizer (AdamW, $\mathrm{lr}\!=\!10^{-4}$, 5-epoch warm-up, cosine schedule), the same training budget (200 epochs of 100 steps each, batch size 32, randomized context length $n_\mathrm{ctx}\!\in\![100,110]$), and the same frozen evaluation sets (500-function validation, 2000-function test) generated once with a fixed seed and loaded by every worker. Two architectural details are scaled with $d$ to keep cross-cell comparisons fair. The FFN hidden width is set to $4d$ rather than fixed at 256, which would silently inflate small models (a fixed-FFN ablation is reported below). The head count is chosen so head dimension is in $[8, 16]$ regardless of $d$, rather than fixed at $4$, which would collapse head dimension at $d\!=\!16$. Latency is measured in ms per batch of 32 on a single GPU after 30 warm-up forward passes. The full grid summary is reported in Tab.~\ref{tab:abl-grid}.

\paragraph{Mechanistic predictions tested.}
Two findings of Secs.~\ref{sec:probing}--\ref{sec:causal} generate concrete, falsifiable predictions about architecture. First, the abrupt $L_1\!\to\!L_2$ emergence reported in Sec.~ \ref{sec:patching} where the causal effect of patching $\bar H$ jumps from 0 to $\approx\!1$ in a single cross-attention step --- predicts that depth beyond two cross-attention layers should yield diminishing returns for spectrally-driven prediction, with the marginal benefit of additional layers attributable to posterior \emph{formatting} rather than spectral \emph{construction}. Second, the subspace concentration of Fig.~\ref{fig:combined_patching}, where a small portion of principal components account for the full causal effect, predicts that the residual-stream dimensionality required to carry the spectral computation is small in absolute terms, although learning to organize that one-dimensional signal in the right direction may demand a larger working budget than the signal itself occupies.

\paragraph{Depth saturates rapidly, consistent with $L_1\!\to\!L_2$ emergence.}
Holding width fixed and varying $L$ (Table ~\ref{tab:abl-grid}), depth gains shrink as $d$ shrinks. At $d\!=\!128$ (MQA), increasing $L$ from 2 to 6 roughly halves the test MSE ($6.7\!\times\!10^{-5}\!\to\!3.1\!\times\!10^{-5}$,
a factor of $2.2$). At $d\!=\!64$ the same $3\!\times$ depth increase yields only $1.4\times$; at $d\!=\!48$ it yields $1.2\times$ and the relationship is not monotone ($L\!=\!3$ outperforms both $L\!=\!2$ and $L\!=\!6$). We read
this as the experimental dual of the manifold-correlation curves of Fig.~\ref{fig:layerwise_manifold}: the first cross-attention layer produces the spectrally-organized $\bar H$, and additional layers chiefly perform downstream formatting --- useful but with diminishing returns, especially when the residual stream is narrow enough that the formatting capacity is already
saturated. The fact that depth helps \emph{more} at $d\!=\!128$ than at
$d\!=\!48$ also supports the formatting interpretation: a wider stream gives
later layers more usable capacity to refine.

\begin{figure}
    \centering
    \includegraphics[width=\linewidth]{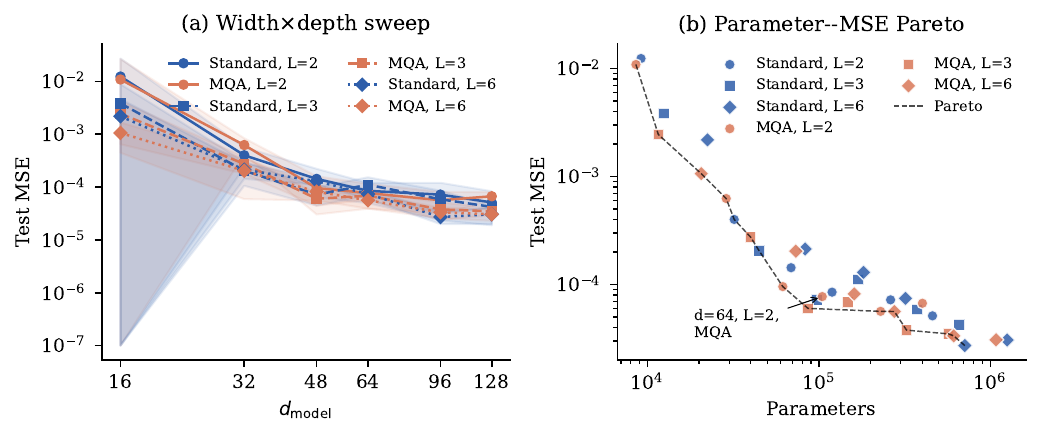}
\caption{Architecture ablation supporting the mechanistic predictions of Secs.~\ref{sec:probing}--\ref{sec:causal}. Test MSE on the held-out 2000-function set across a $6\!\times\!3\!\times\!2$ grid of widths $d$, depths $L$, and attention variants (3 seeds each). \textbf{(a)}~MSE vs.\ $d_{\mathrm{model}}$, with shaded $\pm 1$~std bands across seeds. Width gains plateau by $d\!\approx\!48$--$64$, and the spread between depths shrinks as $d$ shrinks --- consistent with the $L_1\!\to\!L_2$ emergence of spectral coding (Sec.~\ref{sec:patching}): once the first cross-attention step has constructed the spectrally-organized latent, additional layers contribute mainly posterior formatting, with diminishing returns. \textbf{(b)}~Parameter--MSE Pareto frontier (dashed). The annotated configuration ($d\!=\!64$, $L\!=\!2$, MQA) achieves test MSE $7.7\!\times\!10^{-5}$ at $\sim\!105$K parameters and $1.15$~ms/batch --- within $2.5\times$ of the largest configuration ($d\!=\!128$, $L\!=\!6$, Standard; $1.25$M parameters, $3.66$~ms) at $12\times$ fewer parameters and $3.2\times$ lower latency. 12 of 15 Pareto-optimal points use MQA. Full numerical table in Table ~\ref{tab:abl-grid}.}
\label{fig:abl}
\end{figure}

\paragraph{Headline configuration and limitations.}
The most efficient configuration that still approaches the full-capacity performance is $d\!=\!64$, $L\!=\!2$, MQA: test MSE $7.7\!\times\!10^{-5}$, $\sim\!105$K parameters, $1.15$~ms per batch of 32. Our largest configuration ($d\!=\!128$, $L\!=\!6$, Standard) achieves $3.1\!\times\!10^{-5}$ at $\sim\!1.25$M parameters and $3.66$~ms, so the small model trades $2.5\times$ MSE for $12\times$ fewer parameters and $3.2\times$ lower latency. We note that the benchmark is one-dimensional synthetic data drawn from a fixed prior family; the model is a small DVA-PFN, not a production-scale TabPFN.  The ratios reported here should not be ported uncritically to higher dimensions or to TabPFN-scale architectures. Therefore, we present these number in Table \ref{tab:abl-grid} and Figure \ref{fig:abl} as evidence that the mechanistic story has design content, not as a prescription.

\begin{table}[h]
\centering
\small
\caption{Full architecture grid (Appendix~\ref{sec:abl}). Six widths $\times$ three depths $\times$ two attention variants, three seeds each. Mean and standard deviation of test MSE across seeds; latency reported in ms per batch of 32 on a single GPU.}
\label{tab:abl-grid}
\begin{tabular}{rrlrrrr}
\toprule
$d$ & $L$ & attn & params & MSE (mean) & MSE (std) & lat (ms) \\
\midrule
   16 & 2 & MQA      &    8.6K & $1.09\!\times\!10^{-2}$ & $1.58\!\times\!10^{-2}$ & 0.89 \\
   32 & 2 & MQA      &   28.8K & $6.26\!\times\!10^{-4}$ & $2.17\!\times\!10^{-4}$ & 0.80 \\
   48 & 2 & MQA      &   61.6K & $9.57\!\times\!10^{-5}$ & $6.54\!\times\!10^{-5}$ & 0.82 \\
   64 & 2 & MQA      &  104.6K & $7.74\!\times\!10^{-5}$ & $3.83\!\times\!10^{-5}$ & 1.15 \\
   96 & 2 & MQA      &  229.1K & $5.65\!\times\!10^{-5}$ & $2.47\!\times\!10^{-5}$ & 1.06 \\
  128 & 2 & MQA      &  401.6K & $6.71\!\times\!10^{-5}$ & $1.37\!\times\!10^{-5}$ & 1.20 \\
   16 & 3 & MQA      &   11.6K & $2.45\!\times\!10^{-3}$ & $1.80\!\times\!10^{-3}$ & 0.97 \\
   32 & 3 & MQA      &   39.9K & $2.75\!\times\!10^{-4}$ & $9.81\!\times\!10^{-5}$ & 1.20 \\
   48 & 3 & MQA      &   86.4K & $6.02\!\times\!10^{-5}$ & $1.58\!\times\!10^{-5}$ & 1.40 \\
   64 & 3 & MQA      &  147.4K & $6.89\!\times\!10^{-5}$ & $1.09\!\times\!10^{-5}$ & 1.57 \\
   96 & 3 & MQA      &  324.7K & $3.78\!\times\!10^{-5}$ & $1.37\!\times\!10^{-5}$ & 1.58 \\
  128 & 3 & MQA      &  571.1K & $3.49\!\times\!10^{-5}$ & $5.73\!\times\!10^{-6}$ & 1.95 \\
   16 & 6 & MQA      &   20.6K & $1.06\!\times\!10^{-3}$ & $6.17\!\times\!10^{-4}$ & 1.90 \\
   32 & 6 & MQA      &   73.3K & $2.04\!\times\!10^{-4}$ & $1.44\!\times\!10^{-4}$ & 2.44 \\
   48 & 6 & MQA      &  160.7K & $8.20\!\times\!10^{-5}$ & $2.64\!\times\!10^{-5}$ & 2.87 \\
   64 & 6 & MQA      &  275.6K & $5.63\!\times\!10^{-5}$ & $1.71\!\times\!10^{-5}$ & 3.46 \\
   96 & 6 & MQA      &  611.5K & $3.35\!\times\!10^{-5}$ & $7.72\!\times\!10^{-6}$ & 2.61 \\
  128 & 6 & MQA      & 1079.5K & $3.08\!\times\!10^{-5}$ & $7.75\!\times\!10^{-6}$ & 1.93 \\
   16 & 2 & Standard &    9.2K & $1.24\!\times\!10^{-2}$ & $1.48\!\times\!10^{-2}$ & 0.75 \\
   32 & 2 & Standard &   32.1K & $4.01\!\times\!10^{-4}$ & $1.66\!\times\!10^{-4}$ & 0.77 \\
   48 & 2 & Standard &   68.8K & $1.43\!\times\!10^{-4}$ & $7.92\!\times\!10^{-5}$ & 0.97 \\
   64 & 2 & Standard &  119.3K & $8.50\!\times\!10^{-5}$ & $3.52\!\times\!10^{-5}$ & 1.41 \\
   96 & 2 & Standard &  261.9K & $7.21\!\times\!10^{-5}$ & $4.87\!\times\!10^{-5}$ & 1.22 \\
  128 & 2 & Standard &  459.7K & $5.13\!\times\!10^{-5}$ & $3.24\!\times\!10^{-5}$ & 1.38 \\
   16 & 3 & Standard &   12.5K & $3.84\!\times\!10^{-3}$ & $5.05\!\times\!10^{-3}$ & 1.12 \\
   32 & 3 & Standard &   44.8K & $2.05\!\times\!10^{-4}$ & $9.91\!\times\!10^{-5}$ & 1.26 \\
   48 & 3 & Standard &   97.2K & $7.22\!\times\!10^{-5}$ & $2.74\!\times\!10^{-5}$ & 1.33 \\
   64 & 3 & Standard &  169.4K & $1.11\!\times\!10^{-4}$ & $4.26\!\times\!10^{-5}$ & 1.65 \\
   96 & 3 & Standard &  373.9K & $5.88\!\times\!10^{-5}$ & $2.70\!\times\!10^{-5}$ & 1.74 \\
  128 & 3 & Standard &  658.3K & $4.23\!\times\!10^{-5}$ & $2.28\!\times\!10^{-6}$ & 2.28 \\
   16 & 6 & Standard &   22.4K & $2.19\!\times\!10^{-3}$ & $2.47\!\times\!10^{-3}$ & 2.23 \\
   32 & 6 & Standard &   83.1K & $2.14\!\times\!10^{-4}$ & $7.05\!\times\!10^{-5}$ & 2.56 \\
   48 & 6 & Standard &  182.3K & $1.30\!\times\!10^{-4}$ & $1.82\!\times\!10^{-5}$ & 3.12 \\
   64 & 6 & Standard &  319.8K & $7.44\!\times\!10^{-5}$ & $1.50\!\times\!10^{-5}$ & 3.33 \\
   96 & 6 & Standard &  710.0K & $2.72\!\times\!10^{-5}$ & $7.18\!\times\!10^{-6}$ & 3.66 \\
  128 & 6 & Standard & 1253.9K & $3.06\!\times\!10^{-5}$ & $1.06\!\times\!10^{-5}$ & 3.66 \\
\bottomrule
\end{tabular}
\end{table}

\end{document}